\documentclass{article}
\usepackage{ijcai25}

\usepackage{times}
\usepackage{soul}
\usepackage{url}
\usepackage[hidelinks]{hyperref}
\usepackage[utf8]{inputenc}
\usepackage[small]{caption}
\usepackage{graphicx}
\usepackage{amsmath}
\usepackage{amsthm}
\usepackage{booktabs}
\usepackage{algorithm}
\usepackage{algorithmic}
\usepackage[switch]{lineno}

\usepackage{graphicx}

\usepackage{url}            
\usepackage{booktabs}       
\usepackage{multirow}
\usepackage{amsfonts}       
\usepackage{nicefrac}       
\usepackage{microtype}      
\usepackage{makecell} 
\usepackage{float} 

\usepackage{listings} 

\usepackage{dashrule} 
\usepackage{subcaption}
\usepackage{caption}
\usepackage{float}
\usepackage[table]{xcolor} 

\usepackage{multicol}
\usepackage{tocloft} 
\usepackage{appendix} 

\usepackage{tcolorbox} 
\newtcolorbox{simplebox}{
  colback=gray!10,        
  colframe=black,       
  boxrule=0.8pt,        
  arc=4pt,              
  left=0.8em,             
  right=0.8em,            
  top=0.8em,              
  bottom=0.8em,           
  boxsep=0pt,           
}

\newtcolorbox[auto counter, number within=section]{complexbox}[2][]{
  colback=gray!10,        
  colframe=blue!50!black, 
  fonttitle=\bfseries,    
  title=#2,               
  boxrule=0.8pt,          
  arc=4pt,                
  left=1em,               
  right=1em,              
  top=1em,                
  bottom=1em,             
  boxsep=0pt,             
  #1                      
}

\usepackage{threeparttable}

\tcbset{
    commonbox/.style={colframe=white, on line, boxsep=2pt, left=0pt, right=0pt, top=0pt, bottom=0pt},
    graybox/.style={commonbox, colback=gray!30!},
    redbox_lo/.style={commonbox, colback=red!10!},
    redbox_mi/.style={commonbox, colback=red!30!},
    redbox_hi/.style={commonbox, colback=red!50!},
    greenbox_lo/.style={commonbox, colback=green!10!},
    greenbox_mi/.style={commonbox, colback=green!30!},
    greenbox_hi/.style={commonbox, colback=green!50!},
}

\newcommand{\headerthree}[3]{\makecell{\textbf{#1} \\ (#2) \\ #3}}
\newcommand{\headertwo}[2]{\makecell{\textbf{#1} \\ (#2)}}

\newcommand{\vone}[2]{
    #1 
    \ifdim #2 pt > 0 pt 
        \ifdim #2 pt > 2.0 pt
            \tcbox[greenbox_hi]{\scriptsize{($\uparrow$#2)}}
        \else
            \ifdim #2 pt > 0.5 pt
                \tcbox[greenbox_mi]{\scriptsize{($\uparrow$#2)}}
            \else
                \tcbox[greenbox_lo]{\scriptsize{($\uparrow$#2)}}
            \fi
        \fi
    \else
        \ifdim #2 pt < 0 pt 
            \ifdim #2 pt < -3.0 pt
                \tcbox[redbox_hi]{\scriptsize{($\downarrow$\stripminus{#2})}}
            \else
                \ifdim #2 pt < -1.0 pt
                    \tcbox[redbox_mi]{\scriptsize{($\downarrow$\stripminus{#2})}}
                \else
                    \tcbox[redbox_lo]{\scriptsize{($\downarrow$\stripminus{#2})}}
                \fi
            \fi
        \else
            \tcbox[graybox]{\scriptsize{(#2)}}
        \fi
    \fi
}

\newcommand{\vtwo}[2]{
    #1 
    \ifdim #2 pt > 0 pt 
        \ifdim #2 pt > 2.5 pt
            \tcbox[greenbox_hi]{\scriptsize{($\uparrow$#2)}}
        \else
            \ifdim #2 pt > 0.5 pt
                \tcbox[greenbox_mi]{\scriptsize{($\uparrow$#2)}}
            \else
                \tcbox[greenbox_lo]{\scriptsize{($\uparrow$#2)}}
            \fi
        \fi
    \else
        \ifdim #2 pt < 0 pt 
            \ifdim #2 pt < -2.5 pt
                \tcbox[redbox_hi]{\scriptsize{($\downarrow$\stripminus{#2})}}
            \else
                \ifdim #2 pt < -0.5 pt
                    \tcbox[redbox_mi]{\scriptsize{($\downarrow$\stripminus{#2})}}
                \else
                    \tcbox[redbox_lo]{\scriptsize{($\downarrow$\stripminus{#2})}}
                \fi
            \fi
        \else
            \tcbox[graybox]{\scriptsize{(#2)}}
        \fi
    \fi
}

\newcommand\stripminus[1]{\ifdim #1 pt < 0 pt \the\numexpr-#1\else#1\fi}

\usepackage{kotex}

\title{Exploring the Trade-Offs: Quantization Methods, Task Difficulty, and Model Size in Large Language Models From Edge to Giant}

\author{
 Jemin Lee\textsuperscript{1} \and
 Sihyeong Park\textsuperscript{2} \and
 Jinse Kwon\textsuperscript{1}\and
 Jihun Oh\textsuperscript{3} \And
 Yongin Kwon\textsuperscript{1}\thanks{Corresponding author \\ Appendix available at: \url{https://arxiv.org/abs/2409.11055}}
 \affiliations
 \textsuperscript{1}Electronics and Telecommunications Research Institute \\ 
 \textsuperscript{2}Korea Electronics Technology Institute \\
 \textsuperscript{3}Neubla 
\emails
\{leejaymin,kwonse,yongin.kwon\}@etri.re.kr \\
sihyeong@keti.re.kr \and oj9040@gmail.com \\
}

\begin{document}

\definecolor{codegray}{rgb}{0.5,0.5,0.5}
\definecolor{codepurple}{rgb}{0.58,0,0.82}
\definecolor{backcolour}{rgb}{0.95,0.95,0.92}

\lstdefinestyle{mystyle}{
    backgroundcolor=\color{backcolour},   
    commentstyle=\color{codegray},
    keywordstyle=\color{blue},
    numberstyle=\tiny\color{gray},
    stringstyle=\color{codepurple},
    basicstyle=\ttfamily\footnotesize,
    breaklines=true,
    captionpos=b,
    keepspaces=true,
    numbers=left,
    numbersep=5pt,
    showspaces=false,
    showstringspaces=false,
    showtabs=false,
    tabsize=2
}
\lstset{style=mystyle}

\maketitle

\begin{abstract}

Quantization has gained attention as a promising solution for the cost-effective deployment of large and small language models. However, most prior work has been limited to perplexity or basic knowledge tasks and lacks a comprehensive evaluation of recent models like Llama-3.3. 
In this paper, we conduct a comprehensive evaluation of instruction-tuned models spanning 1B to 405B parameters, applying four quantization methods across 13 datasets. 
Our findings reveal that (1) quantized models generally surpass smaller FP16 baselines, yet they often struggle with instruction-following and hallucination detection; (2) FP8 consistently emerges as the most robust option across tasks, and AWQ tends to outperform GPTQ in weight-only quantization; 
(3) smaller models can suffer severe accuracy drops at 4-bit quantization, while 70B-scale models maintain stable performance;
(4) notably, \textit{hard} tasks do not always experience the largest accuracy losses, indicating that quantization magnifies a model’s inherent weaknesses rather than simply correlating with task difficulty; and (5) an LLM-based judge (MT-Bench) highlights significant performance declines in Coding and STEM tasks, though it occasionally reports improvements in reasoning.


\end{abstract}


\captionsetup[figure]{list=no}
\captionsetup[table]{list=no}
\addtocontents{toc}{\protect\setcounter{tocdepth}{0}}
\section{Introduction}

\begin{figure*}[!t]
	\centering
	\includegraphics[width=\linewidth]{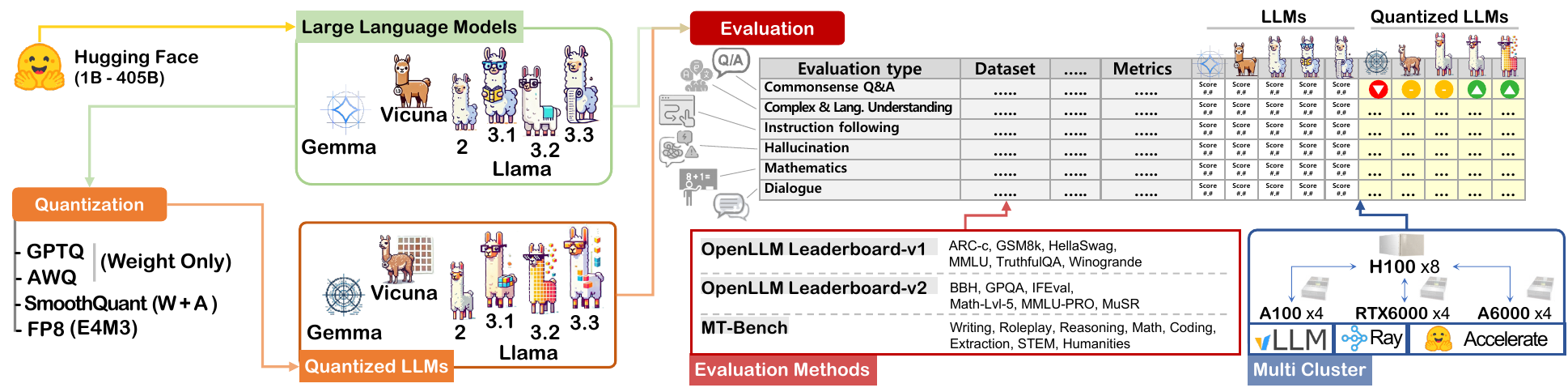}
	\caption{Overview of the evaluation pipeline for quantized LLMs, using a multi-node cluster setup to ensure fast and reliable assessments across multiple benchmarks.
    }
	\label{fig:overview}
\end{figure*}

Despite the remarkable performance of recent large and small language models (LLMs and SLMs), deploying them in resource-constrained environments remains challenging. Even models like Llama-3.3-70B (released in December 2024) and Llama-3.2-1B (released in September 2024) still involve billions of parameters, making them costly to run in both server and mobile-edge scenarios. Low-bit quantization has emerged as a popular solution to reduce the memory and computational overhead of these models. In particular, Post-Training Quantization (PTQ)~\cite{frantar2022optq,lin2024awq,xiao2023smoothquant,micikevicius2022fp8} is widely adopted, as Quantization Aware Training (QAT) often requires extensive retraining~\cite{zhu2023survey,wan2023efficient}.

However, existing research on quantization has largely relied on perplexity-based metrics~\cite{frantar2022optq,lin2024awq,xiao2023smoothquant} and older benchmarks (e.g., ARC, HellaSwag, Winogrande, MMLU)~\cite{yao2023zeroquant,dettmers2023case,liu2023emergent,jin2024comprehensive,li2024evaluating,dutta2024accuracy}, which have become too easy for current models and risk data contamination in recently trained LLMs.
Moreover, more recent architectures like Llama-3.3, Llama-3.2, and Llama-3.1 have not been thoroughly investigated. This gap includes extreme scales, from 1B to over 405B parameters, and omits detailed category-level analysis using LLM-as-judge evaluation methods. Additionally, there has been limited manual inspection to refine and confirm evaluation results~\cite{li2024evaluating,dutta2024accuracy}.

In this paper, we present a comprehensive evaluation of how quantization affects instruction-tuned LLMs. In particular, we aim to address the following research questions (RQs):
\textbf{(RQ1)} Do quantized LLMs outperform smaller original models in most benchmarks, and how do they perform across diverse architectures and small language models (SLMs)? 
\textbf{(RQ2)} How do different quantization methods influence performance across a broad range of tasks, and are there significant differences in how specific approaches (e.g., GPTQ, AWQ, SmoothQuant, FP8) affect task accuracy? 
\textbf{(RQ3)} In what ways do model size and architecture affect the accuracy of quantized models? 
\textbf{(RQ4)} Does higher task difficulty necessarily correlate with greater accuracy degradation under quantization? 
and \textbf{(RQ5)} How does quantization impact the free-form conversation quality of LLMs when evaluated using the MT-Bench framework, which relies on LLMs as judges?

We perform our evaluation in a multi-cluster GPU environment, as illustrated in Figure~\ref{fig:overview}. This setup consists of four servers, each with a distinct GPU configuration, and ensures consistent measurement conditions across all experiments (details in Appendix~\ref{app:reproducibility}). 

Our study applies four quantization methods---GPTQ~\cite{frantar2022optq}, AWQ~\cite{lin2024awq}, SmoothQuant~\cite{xiao2023smoothquant}, and FP8~\cite{micikevicius2022fp8}---to instruction-tuned models ranging from 1B to 405B parameters, including Vicuna~\cite{zheng2023judging}, Gemma~\cite{team2024gemma}, and the Llama family~\cite{dubey2024llama}. We evaluate these models on 13 datasets covering six task categories: commonsense Q\&A, complex knowledge and language understanding, instruction following, hallucination detection, mathematics, and dialogue. Further dataset details are provided in Appendix~\ref{sec:details_datasets} (Table~\ref{tab:benchmarks_summary}).

To compare our results with ongoing community efforts, we synchronize our benchmarks with Huggingface OpenLLM Leaderboard-v1 (covering April 2023 to June 2024) and OpenLLM Leaderboard-v2 (launched on June 26, 2024). However, both versions currently provide only limited data on quantized models, highlighting the need for our comprehensive evaluation. Our key findings are as follows:
\begin{itemize}
    \item Quantized LLMs generally perform better than smaller models on most benchmarks and maintain their advantage across different architectures, showing significant improvements in both large and small language models. However, they still struggle with instruction-following (IFEval) and detecting hallucinations (TruthfulQA).
    
    \item FP8 is the most reliable method for all model sizes and tasks, especially for LLMs with 405B parameters, where SmoothQuant encounters problems. AWQ usually performs better than GPTQ in weight-only quantization, and hardware support makes FP8 even more advantageous.
    
    \item In smaller LLMs, using 4-bit quantization can lead to significant accuracy drops, especially with GPTQ. However, 70B models usually maintain good performance when quantized to 4 bits. While model size is the main factor affecting quantization difficulty, differences in LLM architecture within the same parameter size can also affect accuracy. Nevertheless, AWQ consistently outperforms GPTQ across different tasks and model types.
    
    \item Difficult tasks do not always have the biggest accuracy drops when quantized. The impact depends on the model design and the quantization method used, causing some \textit{hard} tasks to remain stable while some \textit{easy} tasks see bigger decreases. Overall, quantization highlights a model's existing weaknesses, especially in commonsense, logical, or mathematical reasoning.
    
    \item Quantization greatly reduces performance in Coding and STEM tasks, although it sometimes improves reasoning accuracy. Additionally, GPT4-based evaluators can sometimes incorrectly judge wrong answers as correct.
\end{itemize}






\begin{table*}[ht]
    \centering
    \begin{threeparttable}[t]
    \resizebox{.95\textwidth}{!}{
        \begin{tabular}{cccccccccccc}
            \toprule
            \multirow{4.5}{*}{\textbf{Model}} & \multirow{4.5}{*}{\textbf{Method}} & \multirow{4.5}{*}{\textbf{W/A}} & \multirow{4.5}{*}{\makecell{\textbf{Storage} \\ (GB)}} 
                & \multicolumn{6}{c}{\textbf{OpenLLM Leaderboard-v1 $\uparrow$}} \\
            \cmidrule(lr){5-11}
                & & & & \headerthree{ARC-c}{25-shot}{acc\_norm} & \headerthree{GSM8k}{5-shot}{acc} & \headerthree{HellaSwag}{10-shot}{acc\_norm} & \headerthree{MMLU}{5-shot}{acc} & \headerthree{TruthfulQA}{0-shot}{mc2} & \headerthree{Winogrande}{5-shot}{acc} & \textbf{Avg.} \\
            \midrule
            \multirow{3}{*}{Llama-2-7B-Chat} 
                & FP16 & 16 / 16 & 14 & 53.16 & 21.91 & 78.92 & 47.24 & 45.32 & 72.13 & 53.11 \\
                & GPTQ\(^*\) & 4 / 16 & 3.5 & \vone{51.28}{-1.88} & \vone{13.87}{-8.04} & \vone{72.17}{-6.75} & \vone{43.10}{-4.14} & \vone{44.12}{-1.20} & \vone{71.27}{-0.86} & \vone{49.30}{-3.81} \\
                & AWQ & 4 / 16 & 3.5 & \vone{52.47}{-0.69} & \vone{19.63}{-2.28} & \vone{78.13}{-0.79} & \vone{45.34}{-1.90} & \vone{44.28}{-1.04} & \vone{71.19}{-0.94} & \vone{51.84}{-1.27} \\    
            \midrule

            \multirow{3}{*}{Llama-2-13B-Chat} 
                & FP16 & 16 / 16 & 26 & 58.87 & 35.55 & 82.45 & 53.55 & 43.95 & 75.29 & 58.28 \\
                & GPTQ\(^*\) & 4 / 16 & 6.5 & \vone{57.51}{-1.36} & \vone{32.22}{-3.33} & \vone{81.35}{-1.10} & \vone{52.36}{-1.19} & \vone{41.74}{-2.21} & \vone{75.76}{0.47} & \vone{56.82}{-1.46} \\
                & AWQ & 4 / 16 & 6.5 & \vone{57.94}{-0.93} & \vone{34.79}{-0.76} & \vone{81.58}{-0.87} & \vone{53.76}{0.21} & \vone{43.64}{-0.31} & \vone{74.90}{-0.39} & \vone{57.77}{-0.51} \\
            \midrule

            \multirow{3}{*}{Llama-2-70B-Chat} 
                & FP16 & 16 / 16 & 140 & 66.30 & 50.64 & 85.61 & 63.18 & 52.76 & 80.50 & 66.50 \\
                & GPTQ\(^*\) & 4 / 16 & 35 & \vone{62.88}{-3.42} & \vone{50.27}{-0.37} & \vone{84.98}{-0.63} & \vone{61.57}{-1.61} & \vone{51.13}{-1.63} & \vone{79.32}{-1.18} & \vone{65.03}{-1.47} \\
                & AWQ & 4 / 16 & 35 & \vone{65.27}{-1.03} & \vone{48.14}{-2.50} & \vone{85.29}{-0.32} & \vone{62.65}{-0.53} & \vone{52.75}{-0.01} & \vone{79.87}{-0.63} & \vone{65.66}{-0.84} \\
            \midrule

            \multirow{7}{*}{Llama-3.1-8B-it} 
                & FP16 & 16 / 16 & 16 & 60.24 & 76.65 & 80.21 & 68.10 & 54.03 & 76.16 & 69.23 \\
                & FP8 & 8 / 8 & 8 & \vone{61.52}{1.28} & \vone{74.75}{-1.90} & \vone{80.12}{-0.09} & \vone{68.52}{0.42} & \vone{53.81}{-0.22} & \vone{77.43}{1.27} & \vone{69.36}{0.13} \\
                & GPTQ\(^*\) & 4 / 16 & 4 & \vone{61.43}{1.19} & \vone{72.33}{-4.32} & \vone{78.36}{-1.85} & \vone{66.85}{-1.25} & \vone{53.60}{-0.43} & \vone{75.22}{-0.94} & \vone{67.97}{-1.26} \\
                & GPTQ\(^{**}\) & 4 / 16 & 4 & \vone{59.81}{-0.43} & \vone{69.98}{-6.67} & \vone{78.53}{-1.68} & \vone{66.07}{-2.03} & \vone{50.45}{-3.58} & \vone{76.64}{0.48} & \vone{66.91}{-2.32} \\
                & GPTQ\(^{**}\) & 8 / 16 & 8 & \vone{61.01}{0.77} & \vone{75.81}{-0.84} & \vone{80.27}{-0.06} & \vone{68.21}{0.11} & \vone{54.03}{0.00} & \vone{77.19}{1.03} & \vone{69.42}{0.19} \\
                & SmoothQuant & 8 / 8 & 8 & \vone{60.75}{0.51} & \vone{76.12}{-0.53} & \vone{80.08}{-0.13} & \vone{68.22}{0.12} & \vone{53.85}{-0.18} & \vone{77.11}{0.95} & \vone{69.36}{0.13} \\
                & AWQ & 4 / 16 & 4 & \vone{58.53}{-1.71} & \vone{73.39}{-3.26} & \vone{79.10}{-1.11} & \vone{66.26}{-1.84} & \vone{51.87}{-2.16} & \vone{75.37}{-0.79} & \vone{67.42}{-1.81} \\  
            \midrule

            \multirow{7}{*}{Llama-3.1-70B-it} 
                & FP16 & 16 / 16 & 140 & 69.54 & 88.70 & 86.74 & 82.30 & 59.85 & 85.40 & 78.76 \\
                & FP8 & 8 / 8 & 70 & \vone{69.45}{-0.09} & \vone{88.25}{-0.45} & \vone{86.69}{-0.05} & \vone{82.02}{-0.28} & \vone{59.80}{-0.05} & \vone{85.08}{-0.32} & \vone{78.55}{-0.21} \\
                & GPTQ\(^*\) & 4 / 16 & 35 & \vone{69.80}{0.26} & \vone{89.54}{0.84} & \vone{86.28}{-0.46} & \vone{81.40}{-0.90} & \vone{59.37}{-0.48} & \vone{84.69}{-0.71} & \vone{78.51}{-0.25} \\
                & GPTQ\(^{**}\) & 4 / 16 & 35 & \vone{69.97}{0.43} & \vone{89.76}{1.06} & \vone{86.26}{-0.48} & \vone{81.97}{-0.33} & \vone{58.74}{-1.11} & \vone{84.53}{-0.87} & \vone{78.54}{-0.22} \\
                & GPTQ\(^{**}\) & 8 / 16 & 70 & \vone{69.03}{-0.51} & \vone{87.95}{-0.75} & \vone{86.29}{-0.45} & \vone{82.17}{-0.13} & \vone{58.94}{-0.91} & \vone{84.53}{-0.87} & \vone{78.15}{-0.61} \\
                & SmoothQuant & 8 / 8 & 70 & \vone{70.05}{0.51} & \vone{88.55}{-0.15} & \vone{86.56}{-0.18} & \vone{82.10}{-0.20} & \vone{60.39}{0.54} & \vone{85.24}{-0.16} & \vone{78.82}{0.06} \\
                & AWQ & 4 / 16 & 35 & \vone{69.80}{0.26} & \vone{90.83}{2.13} & \vone{86.18}{-0.56} & \vone{81.33}{-0.97} & \vone{59.68}{-0.17} & \vone{84.37}{-1.03} & \vone{78.70}{-0.06} \\
            \midrule
            
            \multirow{5}{*}{Llama-3.1-405B-it} 
                & FP16 & 16 / 16 & 810 & 73.72 & 94.84 & 88.40 & 83.98 & 65.42 & 85.00 & 81.89 \\
                & FP8 & 8 / 8 & 405 & \vone{73.12}{-0.60} & \vone{95.38}{0.54} & \vone{88.32}{-0.08} & \vone{85.91}{1.93} & \vone{64.79}{-0.63} & \vone{85.63}{0.63} & \vone{82.19}{0.30} \\
                & GPTQ\(^{**}\) & 4 / 16 & 202.5 & \vone{72.10}{-1.62} & \vone{94.24}{-0.60} & \vone{88.17}{-0.23} & \vone{85.79}{1.81} & \vone{64.80}{-0.62} & \vone{85.48}{0.48} & \vone{81.76}{-0.13} \\
                & SmoothQuant & 8 / 8 & 405 & \vone{72.01}{-1.71} & \vone{92.72}{-2.12} & \vone{87.53}{-0.87} & \vone{73.28}{-10.70} & \vone{65.19}{-0.23} & \vone{85.95}{0.95} & \vone{79.45}{-2.44} \\
                & AWQ & 4 / 16 & 202.5 & \vone{73.98}{0.26} & \vone{94.84}{0.00} & \vone{88.04}{-0.36} & \vone{85.71}{1.73} & \vone{64.25}{-1.17} & \vone{86.35}{1.35} & \vone{82.20}{0.31} \\
            \midrule
            
            \multirow{5}{*}{Llama-3.2-1B-it} 
                & FP16 & 16 / 16 & 2 & 42.06 & 33.36 & 59.62 & 45.44 & 43.82 & 62.59 & 47.81 \\
                & FP8 & 8 / 8 & 1 & \vone{41.98}{-0.08} & \vone{34.04}{0.68} & \vone{59.20}{-0.42} & \vone{45.38}{-0.06} & \vone{43.18}{-0.64} & \vone{62.67}{0.08} & \vone{47.74}{-0.07} \\
                & GPTQ$^{***}$ & 4 / 16 & 0.5 & \vone{34.90}{-7.16} & \vone{8.04}{-25.32} & \vone{43.14}{-16.48} & \vone{40.20}{-5.24} & \vone{41.92}{-1.90} & \vone{57.85}{-4.74} & \vone{37.67}{-10.14} \\
                & SmoothQuant & 8 / 8 & 0.5 & \vone{42.06}{0.00} & \vone{33.13}{-0.23} & \vone{59.51}{-0.11} & \vone{45.19}{-0.25} & \vone{43.40}{-0.42} & \vone{61.48}{-1.11} & \vone{47.46}{-0.35} \\
                & AWQ & 4 / 16 & 0.5 & \vone{38.48}{-3.58} & \vone{17.97}{-15.39} & \vone{53.82}{-5.80} & \vone{40.51}{-4.39} & \vone{42.61}{-1.21} & \vone{59.12}{-3.47} & \vone{42.09}{-5.72} \\
            \midrule
            
            \multirow{5}{*}{Llama-3.2-3B-it} 
                & FP16 & 16 / 16 & 6 & 52.13 & 64.52 & 73.08 & 59.59 & 47.79 & 69.61 & 61.45 \\
                & FP8 & 8 / 8 & 3 & \vone{51.37}{-0.76} & \vone{63.31}{-1.21} & \vone{73.04}{-0.04} & \vone{59.74}{0.15} & \vone{49.98}{0.19} & \vone{69.14}{-0.47} & \vone{61.10}{-0.35} \\
                & GPTQ$^{***}$ & 4 / 16 & 1.5 & \vone{50.26}{-1.87} & \vone{60.20}{-4.32} & \vone{71.19}{-1.89} & \vone{57.90}{-1.69} & \vone{49.52}{-0.27} & \vone{68.75}{-0.86} & \vone{59.64}{-1.81} \\
                & SmoothQuant & 8 / 8 & 1.5 & \vone{51.37}{-0.76} & \vone{63.76}{-0.76} & \vone{72.61}{-0.47} & \vone{56.69}{0.10} & \vone{49.72}{-0.07} & \vone{69.61}{0.00} & \vone{61.13}{-0.32} \\
                & AWQ & 4 / 16 & 1.5 & \vone{50.51}{-1.62} & \vone{61.41}{-3.11} & \vone{71.27}{-1.81} & \vone{58.94}{-0.65} & \vone{49.03}{-0.76} & \vone{67.4}{-2.21} & \vone{59.76}{-1.69} \\
            \midrule

            \multirow{3}{*}{Llama-3.3-70B-it} 
                & FP16 & 16 / 16 & 140 & 71.67 & 90.83 & 86.39 & 82.20 & 60.90 & 83.98 & 79.33 \\
                & GPTQ$^{***}$ & 4 / 16 & 35 & \vone{69.71}{-1.96} & \vone{89.39}{-1.44} & \vone{85.58}{-0.81} & \vone{81.63}{-0.57} & \vone{61.25}{0.35} & \vone{84.21}{0.23} & \vone{78.63}{-0.70} \\
                & AWQ & 4 / 16 & 35 & \vone{70.82}{-0.85} & \vone{88.17}{-2.66} & \vone{85.73}{-0.66} & \vone{81.45}{-0.75} & \vone{60.82}{-0.08} & \vone{83.98}{0.00} & \vone{78.50}{-0.83} \\
            \bottomrule

        \end{tabular}
    }
    \end{threeparttable}%
    \caption{Evaluation of Llama families on OpenLLM Leaderboard-v1. \(^*\), \(^{**}\), and \(^{***}\) denote the use of AutoGPTQ, llmcompressor, and AutoRound for GPTQ quantization, respectively.}
    \label{tab:evaluation_results}
\end{table*}

\begin{table*}[ht]
    \centering
    \begin{threeparttable}[t]
    \resizebox{.95\textwidth}{!}{
        \begin{tabular}{cccccccccccc}
            \toprule
            \multirow{3.5}{*}{\textbf{Model}} & \multirow{3.5}{*}{\textbf{Method}} & \multirow{3.5}{*}{\textbf{W/A}} & \multirow{3.5}{*}{\makecell{\textbf{Storage} \\ (GB)}} 
                & \multicolumn{6}{c}{\textbf{OpenLLM Leaderboard-v2 $\uparrow$}} \\
            \cmidrule(lr){5-11}
                & & & & \headertwo{BBH}{3-shot} & \headertwo{GPQA}{0-shot} & \headertwo{IFEval}{0-shot} & \headertwo{Math-Lvl-5}{4-shot} & \headertwo{MMLU-PRO}{5-shot} & \headertwo{MuSR}{0-shot} & \textbf{Avg.} \\
            \midrule
            \multirow{3}{*}{Llama-2-7B-Chat} 
                & FP16 & 16 / 16 & 14 & 12.23 & 1.59 & 35.31 & 1.93 & 11.04 & 8.89 & 11.83 \\
                & GPTQ\(^*\) & 4 / 16 & 3.5 & \vtwo{7.96}{-4.27} & \vtwo{0.85}{-0.74} & \vtwo{30.59}{-4.72} & \vtwo{1.43}{-0.50} & \vtwo{7.58}{-3.46} & \vtwo{3.73}{-5.16} & \vtwo{8.69}{-3.14} \\
                & AWQ & 4 / 16 & 3.5 & \vtwo{10.78}{-1.45} & \vtwo{3.54}{1.95} & \vtwo{31.57}{-3.74} & \vtwo{1.81}{-0.12} & \vtwo{7.75}{-3.29} & \vtwo{10.62}{1.73} & \vtwo{11.01}{-0.82} \\
            \midrule

            \multirow{3}{*}{Llama-2-13B-Chat} 
                & FP16 & 16 / 16 & 26 & 16.87 & 4.03 & 37.45 & 1.56 & 15.36 & 10.00 & 14.21 \\
                & GPTQ\(^*\) & 4 / 16 & 6.5 & \vtwo{16.92}{0.05} & \vtwo{4.27}{0.24} & \vtwo{33.40}{-4.05} & \vtwo{2.21}{0.65} & \vtwo{15.44}{0.08} & \vtwo{8.30}{-1.70} & \vtwo{13.42}{-0.79} \\
                & AWQ & 4 / 16 & 6.5 & \vtwo{16.54}{-0.33} & \vtwo{6.96}{2.93} & \vtwo{33.28}{-4.17} & \vtwo{1.80}{0.24} & \vtwo{15.57}{0.21} & \vtwo{10.39}{0.39} & \vtwo{14.09}{-0.12} \\
            \midrule

            \multirow{3}{*}{Llama-2-70B-Chat} 
                & FP16 & 16 / 16 & 140 & 29.42 & 6.72 & 44.11 & 2.64 & 25.01 & 6.32 & 19.04 \\
                & GPTQ\(^*\) & 4 / 16 & 35 & \vtwo{25.80}{-3.62} & \vtwo{5.25}{-1.47} & \vtwo{41.52}{-2.59} & \vtwo{3.27}{0.63} & \vtwo{23.87}{-1.14} & \vtwo{7.06}{0.74} & \vtwo{17.80}{-1.24} \\
                & AWQ & 4 / 16 & 35 & \vtwo{28.63}{-0.79} & \vtwo{6.72}{0.00} & \vtwo{42.89}{-1.22} & \vtwo{1.95}{-0.69} & \vtwo{24.44}{-0.57} & \vtwo{5.56}{-0.76} & \vtwo{18.37}{-0.67} \\
            \midrule

            \multirow{7}{*}{Llama-3.1-8B-it} 
                & FP16 & 16 / 16 & 16 & 30.11 & 6.23 & 50.09 & 11.69 & 30.90 & 8.88 & 22.98 \\
                & FP8 & 8 / 8 & 8 & \vtwo{29.20}{-0.91} & \vtwo{5.49}{-0.74} & \vtwo{49.16}{-0.93} & \vtwo{12.01}{0.32} & \vtwo{30.92}{0.02} & \vtwo{6.95}{-1.93} & \vtwo{22.29}{-0.69} \\
                & GPTQ\(^*\) & 4 / 16 & 4 & \vtwo{25.86}{-4.25} & \vtwo{7.20}{0.97} & \vtwo{47.95}{-2.14} & \vtwo{9.49}{-2.20} & \vtwo{29.60}{-1.30} & \vtwo{6.03}{-2.85} & \vtwo{21.02}{-1.96} \\
                & GPTQ\(^{**}\) & 4 / 16 & 4 & \vtwo{25.83}{-4.28} & \vtwo{6.72}{0.49} & \vtwo{44.81}{-5.28} & \vtwo{8.85}{-2.84} & \vtwo{28.16}{-2.74} & \vtwo{10.58}{1.70} & \vtwo{20.83}{-2.15} \\
                & GPTQ\(^{**}\) & 8 / 16 & 8 & \vtwo{29.97}{-0.14} & \vtwo{6.23}{0.00} & \vtwo{50.53}{0.44} & \vtwo{11.94}{0.25} & \vtwo{31.19}{0.29} & \vtwo{7.80}{-1.08} & \vtwo{22.94}{-0.04} \\
                & SmoothQuant & 8 / 8 & 8 & \vtwo{30.19}{0.08} & \vtwo{2.56}{-3.67} & \vtwo{50.25}{0.16} & \vtwo{12.77}{1.08} & \vtwo{30.75}{-0.15} & \vtwo{8.12}{-0.76} & \vtwo{22.44}{-0.54} \\
                & AWQ & 4 / 16 & 4 & \vtwo{25.73}{-4.38} & \vtwo{5.98}{-0.25} & \vtwo{47.97}{-2.12} & \vtwo{10.02}{-1.67} & \vtwo{29.08}{-1.82} & \vtwo{6.74}{-2.14} & \vtwo{20.92}{-2.06} \\
            \midrule
            
            \multirow{7}{*}{Llama-3.1-70B-it} 
                & FP16 & 16 / 16 & 140 & 55.90 & 16.48 & 75.48 & 28.68 & 48.00 & 19.32 & 40.64 \\
                & FP8 & 8 / 8 & 70 & \vtwo{55.54}{-0.36} & \vtwo{16.24}{-0.24} & \vtwo{75.75}{0.27} & \vtwo{28.92}{0.24} & \vtwo{47.84}{-0.16} & \vtwo{19.35}{0.03} & \vtwo{40.61}{-0.03} \\
                & GPTQ\(^*\) & 4 / 16 & 35 & \vtwo{53.65}{-2.25} & \vtwo{17.70}{1.22} & \vtwo{73.26}{-2.22} & \vtwo{27.26}{-1.42} & \vtwo{47.49}{-0.51} & \vtwo{20.33}{1.01} & \vtwo{39.95}{-0.69} \\
                & GPTQ\(^{**}\) & 4 / 16 & 35 & \vtwo{55.79}{-0.11} & \vtwo{14.04}{-2.44} & \vtwo{72.71}{-2.77} & \vtwo{26.16}{-2.52} & \vtwo{46.97}{-1.03} & \vtwo{16.93}{-2.39} & \vtwo{38.77}{-1.87} \\
                & GPTQ\(^{**}\) & 8 / 16 & 70 & \vtwo{54.79}{-1.11} & \vtwo{2.81}{-13.67} & \vtwo{66.66}{-8.82} & \vtwo{29.06}{0.38} & \vtwo{47.56}{-0.44} & \vtwo{20.42}{1.10} & \vtwo{36.88}{-3.76} \\
                & SmoothQuant & 8 / 8 & 70 & \vtwo{55.06}{-0.84} & \vtwo{16.24}{-0.24} & \vtwo{74.78}{-0.70} & \vtwo{27.90}{-0.78} & \vtwo{47.20}{-0.80} & \vtwo{20.32}{1.00} & \vtwo{40.25}{-0.39} \\
                & AWQ & 4 / 16 & 35 & \vtwo{54.08}{-1.82} & \vtwo{16.48}{0.00} & \vtwo{75.15}{-0.33} & \vtwo{27.85}{-0.83} & \vtwo{47.07}{-0.93} & \vtwo{21.69}{2.37} & \vtwo{40.39}{-0.25} \\
            \midrule
            
            \multirow{5}{*}{Llama-3.1-405B-it} 
                & FP16 & 16 / 16 & 810 & 66.81 & 26.25 & 76.18 & 37.06 & 60.01 & 19.86 & 47.70  \\
                & FP8 & 8 / 8 & 405 & \vtwo{65.22}{-1.59} & \vtwo{27.37}{1.22} & \vtwo{72.44}{-3.74} & \vtwo{35.86}{-1.20} & \vtwo{59.67}{-0.34} & \vtwo{17.83}{-2.03} & \vtwo{46.42}{-1.28} \\
                & GPTQ\(^{**}\) & 4 / 16 & 202.5 & \vtwo{66.21}{-0.60} & \vtwo{23.57}{-2.68} & \vtwo{72.90}{-3.28} & \vtwo{34.74}{-2.32} & \vtwo{59.11}{-0.90} & \vtwo{18.92}{-0.94} & \vtwo{45.91}{-1.79} \\
                & SmoothQuant & 8 / 8 & 405 & \vtwo{54.46}{-12.35} & \vtwo{16.73}{-9.52} & \vtwo{70.34}{-5.84} & \vtwo{35.01}{-2.05} & \vtwo{18.24}{-41.77} & \vtwo{18.60}{-1.26} & \vtwo{35.56}{-12.14} \\
                & AWQ & 4 / 16 & 202.5 & \vtwo{65.50}{-1.31} & \vtwo{26.50}{0.25} & \vtwo{47.52}{-28.66} & \vtwo{38.13}{1.07} & \vtwo{58.63}{-1.38} & \vtwo{19.69}{-0.17} & \vtwo{42.66}{-5.04} \\
            \midrule
            
            \multirow{5}{*}{Llama-3.2-1B-it} 
                & FP16 & 16 / 16 & 2 & 8.32 & 1.34 & 41.61 & 4.20 & 10.60 & 3.70 & 11.63 \\
                & FP8 & 8 / 8 & 1 & \vtwo{8.83}{0.51} & \vtwo{1.59}{0.25} & \vtwo{43.18}{1.57} & \vtwo{3.51}{-0.69} & \vtwo{10.28}{-0.32} & \vtwo{3.49}{-0.21} & \vtwo{11.81}{0.18} \\
                & GPTQ$^{***}$ & 4 / 16 & 0.5 & \vtwo{3.98}{-4.34} & \vtwo{0.85}{-0.49} & \vtwo{25.60}{-16.01} & \vtwo{0.30}{-3.90} & \vtwo{6.72}{-3.88} & \vtwo{1.86}{-1.84} & \vtwo{6.55}{-5.08} \\
                & SmoothQuant & 8 / 8 & 0.5 & \vtwo{8.22}{-0.10} & \vtwo{4.03}{2.69} & \vtwo{41.68}{0.07} & \vtwo{3.25}{-0.95} & \vtwo{10.17}{-0.43} & \vtwo{3.27}{-0.43} & \vtwo{11.77}{0.14} \\
                & AWQ & 4 / 16 & 0.5 & \vtwo{5.21}{-3.11} & \vtwo{4.27}{2.93} & \vtwo{31.21}{-10.40} & \vtwo{1.56}{-2.64} & \vtwo{7.61}{-2.99} & \vtwo{2.84}{-0.86} & \vtwo{8.78}{-2.85} \\
            \midrule

            \multirow{5}{*}{Llama-3.2-3B-it} 
                & FP16 & 16 / 16 & 6 & 20.77 & 7.69 & 53.56 & 11.20 & 22.24 & 6.92 & 20.40 \\
                & FP8 & 8 / 8 & 3 & \vtwo{21.00}{0.23} & \vtwo{7.69}{0.00} & \vtwo{53.32}{-0.24} & \vtwo{9.78}{-1.42} & \vtwo{21.99}{-0.25} & \vtwo{7.30}{0.38} & \vtwo{20.18}{-0.22} \\
                & GPTQ$^{***}$ & 4 / 16 & 1.5 & \vtwo{18.82}{-1.95} & \vtwo{5.98}{-1.71} & \vtwo{52.81}{-0.75} & \vtwo{8.05}{-3.14} & \vtwo{19.31}{-2.93} & \vtwo{4.19}{-2.73} & \vtwo{18.19}{-2.21} \\
                & SmoothQuant & 8 / 8 & 1.5 & \vtwo{21.01}{0.24} & \vtwo{9.65}{-1.96} & \vtwo{53.96}{0.40} & \vtwo{9.58}{-1.62} & \vtwo{21.83}{-0.41} & \vtwo{6.07}{-0.85} & \vtwo{20.35}{-0.05} \\
                & AWQ & 4 / 16 & 1.5 & \vtwo{19.45}{-1.32} & \vtwo{8.18}{0.49} & \vtwo{51.5}{-2.06} & \vtwo{8.3}{-2.9} & \vtwo{20.79}{-1.45} & \vtwo{7.19}{-0.27} & \vtwo{19.23}{-1.17} \\
            \midrule

            \multirow{3}{*}{Llama-3.3-70B-it} 
                & FP16 & 16 / 16 & 140 & 56.78 & 30.40 & 69.03 & 30.95 & 49.78 & 22.28 & 43.20 \\
                & GPTQ$^{***}$ & 4 / 16 & 35 & \vtwo{52.48}{-4.31} & \vtwo{28.94}{-1.46} & \vtwo{65.71}{-3.32} & \vtwo{28.55}{-2.40} & \vtwo{48.24}{-1.54} & \vtwo{21.38}{-0.90} & \vtwo{40.88}{-2.32} \\
                & AWQ & 4 / 16 & 35 & \vtwo{55.89}{-0.89} & \vtwo{28.69}{-1.71} & \vtwo{69.61}{0.58} & \vtwo{29.42}{-1.53} & \vtwo{48.57}{-1.21} & \vtwo{20.67}{-1.61} & \vtwo{42.14}{-1.06} \\
            \bottomrule

        \end{tabular}
    }
    \end{threeparttable}%
    \caption{Evaluation of Llama families on OpenLLM Leaderboard-v2. \(^*\), \(^{**}\), and \(^{***}\) denote the use of AutoGPTQ, llmcompressor, and AutoRound for GPTQ quantization, respectively.}
    \label{tab:evaluation_results_v2}
\end{table*}

\section{Related Work}
\textbf{Quantization for LLMs}. There are two main types of quantization methods for LLMs: post-training quantization (PTQ) and quantization-aware training (QAT). Due to the size and training complexity of LLMs, QAT is challenging to apply, and as a result, only limited research has been conducted in this area. Consequently, the majority of quantization research for LLMs has focused on PTQ approaches~\cite{zhu2023survey,wan2023efficient}.

LLM.int8()~\cite{dettmers2022gpt3} is a post-training quantization method that uses 8-bit weights and activations to reduce the memory footprint of large models while maintaining performance. This dynamically adapts to ensure sensitive components of the computation retain higher precision when needed.
GPTQ~\cite{frantar2022optq} is a layer-wise quantization that uses inverse Hessian information to reduce the number of bits per weight while maintaining low accuracy loss.
AWQ~\cite{lin2024awq} proposed that preserving a small portion of important weights is a key part of reducing quantization errors. As part of an activation-aware strategy, AWQ focused on channels with larger activation magnitudes and used per-channel scaling.
SmoothQuant~\cite{xiao2023smoothquant} is a method that smooths activation outliers before quantization, improving robustness in large-scale models and enabling more effective 8-bit quantization.
Outlier Suppression+~\cite{osplus} reduces the impact of extreme outliers in activations, allowing for more efficient quantization by normalizing problematic values without degrading model accuracy.
QLoRA~\cite{dettmers2024qlora} combines low-rank adaptation with quantization to achieve efficient fine-tuning of large models while minimizing computational costs and memory usage.

However, these quantization algorithm works have been evaluated only on basic datasets such as perplexity, ARC, and MMLU, which were released 2-3 years ago, and they do not sufficiently take into account the recent advancements in LLMs and SLMs. Therefore, for a safe application of quantization in LLM services, a more comprehensive performance analysis is necessary.

\noindent\textbf{Evaluating LLMs.}
Several studies have explored the effects of model quantization on the performance of LLMs, focusing on various aspects. For instance, Yao et al.~\cite{yao2023zeroquant} investigated the impact of quantization on both weights and activations in language modeling tasks. In contrast, Liu et al.~\cite{liu2023emergent} concentrated solely on evaluating three emergent abilities of quantized LLMs, neglecting crucial tasks such as trustworthiness, dialogue, and long-context processing.

Hong et al.~\cite{hong2024decoding} expanded the scope by examining trustworthiness dimensions in the assessment of LLM compression techniques. However, most studies have predominantly relied on accuracy as the primary evaluation metric, with limited attention paid to alternative metrics. For example, Zhang et al.~\cite{zhang2024llmeval} proposed additional evaluation metrics, including fluency, informativeness, coherence, and harmlessness, alongside accuracy.

Efforts to establish more comprehensive evaluation benchmarks have also been made. Jaiswal et al.~\cite{jaiswal2023compressing} developed a benchmark from existing datasets to evaluate compressed models, while Li et al.~\cite{li2024evaluating} and Jin et al.~\cite{jin2024comprehensive} assessed various quantization techniques across different tasks. Namburi et al.~\cite{namburi2023cost} explored how compression and pruning affect the parametric knowledge of LLMs.

Dutta et al.~\cite{dutta2024accuracy} proposed a new metric from the perspective of “Flip” errors, emphasizing the importance of evaluating model robustness by accounting for inconsistencies and reversals in predictions, thereby going beyond traditional accuracy-focused metrics. Kurtic et al.~\cite{kurtic2024give} investigated the accuracy-performance trade-offs in quantizing LLMs, evaluating formats like FP8, INT8, and INT4 across various tasks and proposing practical guidelines for efficient LLM deployment at different model scales.

Xu et al.~\cite{xu2024beyond} explored the challenges of evaluating multilingual LLMs across diverse languages and cultures, emphasizing the development of culturally and linguistically inclusive benchmarks for fair evaluation. Liu et al.~\cite{liu2024evaluating} provided a comprehensive examination of the generalization ability, focusing on their performance across various tasks and datasets. This study highlights the importance of designing benchmarks and metrics that accurately reflect real-world applications while identifying the limitations of current evaluation strategies.


To our knowledge, no prior study has comprehensively examined the effects of quantization across a wide range of model sizes—from 1B to 405B parameters—encompassing both SLMs and LLMs, including the latest architectures such as Llama-3.1, Llama-3.2, and Llama-3.3. Furthermore, existing research has not conducted detailed category-level analyses through cross-architecture comparisons or employed manual inspection using LLM-as-judge qualitative methods. Additionally, no work has compared the trends observed in LLM-as-judge (MT-Bench) evaluations with leaderboard results to identify new trends in quantization impacts.

\section{Evaluation Procedure}
To handle LLMs, which cannot be processed on a single server, and to ensure fast and reliable evaluations, we developed a structured evaluation pipeline based on a multi-node cluster setup. Figure~\ref{fig:overview} presents an overview of the implemented pipeline for evaluating quantized LLMs. 
The evaluated LLMs include the Vicuna, Gemma, and Llama families, ranging in size from 1B to 405B. Each model is quantized using GPTQ, AWQ, SmoothQuant, and FP8 methods. 
The evaluation is conducted using lm-eval and MT-Bench as benchmarking tools.
The multi-node cluster used for evaluation is implemented with vLLM and consists of four servers: H100-80Gx8, A100-80Gx4, RTX 6000-48Gx4, and A6000-48Gx4. Additionally, the Huggingface library is integrated into the pipeline to support model hosting and benchmarking. The evaluation is distributed across a multi-cluster environment to ensure a thorough performance assessment. 
If vLLM cannot be used for processing, we used the Huggingface Accelerate library instead, which is slower but shows better comparability.
The versions of all tools used are provided in Appendix~\ref{app:random_factor}.

\section{Experimental Setup} 

\subsection{Datasets} 
We conducted a comprehensive evaluation of the quantized LLMs on widely adopted benchmarks, grouped into six main categories: CommonSenseQA (ARC, HellaSwag, Winogrande), Knowledge and Language Understanding (MMLU, GPQA, MMLU-PRO, BBH, MuSR), Instruction Following (IFEval), Hallucination (TruthfulQA), Mathematics (GSM8K, MATH-Lvl-5), and Dialogue (MT-Bench). 
Additional information about these datasets can be found in Appendix~\ref{sec:details_datasets}, and an overview of all benchmarks is provided in Table~\ref{tab:benchmarks_summary} (Appendix).

\subsection{Reproducibility Details}
Both OpenLLM Leaderboard V1 and V2 follow the same methodology outlined on the HuggingFace Leaderboard's page, including identical normalization procedures. Additional information on leaderboard calculations is provided in Appendix~\ref{app:details_leaderboard_v1_v2}.

We used the \textit{greedy decoding} strategy to maintain deterministic output tokens across runs.
The detailed configuration for the \textit{greedy decoding} strategy is presented in Appendix~\ref{app:random_factor}, which also lists the specific versions of each package used. Furthermore, we record all random seeds for Python, NumPy, Torch, and few-shot setups in Appendix~\ref{app:random_factor}, ensuring the reproducibility of our experimental results.

\subsection{Quantization Methods and Calibration Data}
We evaluated multiple PTQ methods, including GPTQ, AWQ, SmoothQuant, and FP8. 
GPTQ and AWQ focus on weight-only quantization, while SmoothQuant and FP8 apply to both weights and activations. 
For a detailed overview of each quantization method, refer to Appendix~\ref{app:quant_method}, with configuration details and group sizes described in Appendix~\ref{app:quant_config}.

The selection and configuration of calibration datasets are crucial for maintaining consistent performance across models. We used default settings for the number of samples and sequence lengths, as detailed in Appendix~\ref{app:calibration_dataset} and summarized in Table~\ref{tab:quant_calibration}. These settings may vary depending on the specific algorithms and tools but generally ensure stable results for our experiments.

\subsection{Models} 
We applied quantization techniques to 12 instruction-tuned open LLMs, including the Vicuna~\cite{zheng2023judging}, Gemma~\cite{team2024gemma}, and Llama~\cite{dubey2024llama} families, with model sizes ranging from 1B to 405B. These models were released between June 2023 and December 2024 and were downloaded from HuggingFace's model sources.

All models were evaluated using 13 benchmark datasets, applying GPTQ, AWQ, SmoothQuant, and FP8 quantization. However, due to runtime limitations (over 30 days), we did not measure the original model accuracy or test all GPTQ configurations for Llama-3.1-405B.
Also, due to space limitations, the experimental results for the Vicuna and Gemma models are provided in the Appendix~\ref{app:additional_models}

\section{Experimental Results}
This section presents experimental results addressing five research questions, detailing the performance impact of quantization across 13 datasets for three model families of varying sizes. Table~\ref{tab:evaluation_results} and Table ~\ref{tab:evaluation_results_v2} summarize the experimental results from OpenLLM Leaderboard-v1 and v2, respectively.

\subsection{RQ1: Do quantized LLMs outperform smaller original models on most benchmarks, and how do they fare across different architectures and SLMs?}

We observe that quantized LLMs generally outperform smaller, uncompressed models across a wide range of benchmarks. For instance, a 4-bit Llama-2-13B (6.5 GB) outperforms an FP16 Llama-2-7B (14 GB) on most tasks, despite its reduced size. However, in TruthfulQA (hallucination testing) and IFEval (instruction-following), the FP16 Llama-2-7B still performs better, indicating that quantization can compromise alignment and adherence to instructions.

Similarly, quantizing Llama-3.1-405B to 4 bits (202.5 GB) yields higher accuracy than the FP16 Llama-3.1-70B (140 GB) across various tasks; yet, the instruction-following IFEval benchmark again highlights a shortfall in the quantized model. This performance gap holds across different model architectures: although Llama-3.3-70B demonstrates improvements over Llama-3.1-70B, a 4-bit Llama-3.1-405B can still outperform the uncompressed Llama-3.3-70B.

In edge-focused SLMs, quantization produces significantly larger improvements. For example, quantizing Llama-3.2-3B (SmoothQuant) improves accuracy by 13.32\% on OpenLLM Leaderboard-v1 and 8.72\% on OpenLLM Leaderboard-v2 compared to the FP16-Llama-3.2-1B. Such improvements exceed the margin typically observed in larger models (7B to 405B).

\begin{simplebox}
\textbf{RQ1 Findings:} Quantized LLMs consistently outperform smaller models across most benchmarks and maintain this advantage across different architectures, with significant gains observed in both large models and edge-focused SLMs. However, tasks like instruction-following (IFEval) and hallucination detection (TruthfulQA) remain challenging for quantized models.
\end{simplebox}

\subsection{RQ2: How do different quantization methods affect the performance of models across diverse tasks? Are there noticeable differences in how specific methods (e.g., GPTQ, AWQ, SmoothQuant, FP8) impact task accuracy?}

In most cases, weight-only quantization methods (GPTQ and AWQ) and activation quantization methods (SmoothQuant and FP8) exhibit similar performance. However, for Llama-3.1-405B, SmoothQuant's activation quantization resulted in a significant accuracy drop compared to other methods, with an average decrease of up to 10.86\% compared to FP8 on the OpenLLM Leaderboard-v2 datasets. This decline occurs because SmoothQuant was originally designed to handle the high activation ranges observed in models up to the size of OPT-175B. Consequently, at the 405B scale of Llama-3.1, the algorithm likely did not account for certain factors that are critical at this larger scale. 
In contrast, for smaller models, such as SLMs with 1B and 3B parameters, the average accuracy drop remains below 1\%.

When comparing weight-only quantization methods, AWQ consistently outperforms GPTQ across various LLMs on overall benchmark scores. Additionally, different implementations of GPTQ, such as AutoGPTQ and llmcompressor, demonstrate notable performance differences, with the oldest GPTQ implementation library, AutoGPTQ, still maintaining stable and consistent performance.

When both weight and activation quantization are required at 8 bits, FP8 proves to be highly effective, even on challenging tasks from the OpenLLM Leaderboard-v2. This effectiveness spans models of all sizes, from the largest, such as Llama-3.1-405B, to the smallest, like Llama-3.2-1B. Therefore, FP8 offers greater stability compared to SmoothQuant and is advantageous to use when supported by hardware such as the NVIDIA H100 GPU and RTX 6000 Ada, as it provides benefits in both latency and throughput. Additionally, applying FP8 to both weights and activations allows for a reduction of the KV cache size by half, which is highly beneficial during LLM decoding phases where I/O bottlenecks are a concern. This FP8 KV cache feature is supported by vLLM.

\begin{simplebox}
\textbf{RQ2 Findings:} FP8 is the most stable option across all model sizes and tasks, particularly in large LLMs where SmoothQuant performs poorly, whereas AWQ regularly outperforms GPTQ in weight-only quantization, and specialized hardware enhances FP8's advantages.
\end{simplebox}

\subsection{RQ3: How do model size and architecture influence the accuracy of quantized models?}
We evaluated GPTQ across 13 datasets and observed that its accuracy can degrade significantly under 4-bit quantization---a behavior not reflected in perplexity-based evaluations alone. For smaller models, such as Llama-3.2-1B, 4-bit quantization causes particularly severe accuracy drops (e.g., -25.32\% on GSM8k and -16.01\% on IFEval). These declines tend to be more pronounced in GPTQ than in AWQ, suggesting that SmoothQuant or FP8 at 8-bit may be necessary to maintain accuracy for 1B-scale models.

With mid-sized models like Llama-3.1-8B (GPTQ-8bit), we noted average accuracy improvements of +2.51\% and +2.11\% over 4-bit on OpenLLM Leaderboard-v1 and OpenLLM Leaderboard-v2, respectively. However, at the 70B scale, 4-bit outperformed 8-bit by +1.89\%, indicating a reverse trend for larger models. To examine architectural differences at a fixed scale, we compared Llama-2, Llama-3.1, and Llama-3.3 at the 70B size and found that AWQ consistently surpassed GPTQ, delivering stable accuracy even at 4-bit quantization.

\begin{simplebox}
\textbf{RQ3 Findings:} 
In smaller LLMs, 4-bit quantization often leads to significant accuracy loss (especially with GPTQ), whereas 70B-scale models can maintain stable performance with 4-bit. Although size primarily drives quantization difficulty, architectural variations within the same scale can also influence accuracy. Nonetheless, AWQ consistently outperforms GPTQ across diverse tasks and model families.
\end{simplebox}

\subsection{RQ4: Does higher task difficulty always correlate with greater accuracy degradation under quantization?}

Contrary to common assumptions, tasks widely considered challenging (e.g., GSM8K, MMLU, Math-Lvl-5) did not consistently show the greatest performance drops when quantized. For instance, among 12 tested models quantized with AWQ, MMLU accuracy remained largely unchanged, with some models even exhibiting slight improvements. In contrast, tasks generally regarded as less knowledge-intensive (ARC-challenge, TruthfulQA, Winogrande) occasionally experienced declines exceeding -3\%. 

Rather than strictly depending on task difficulty, quantization appears to magnify existing weaknesses in a model’s ability to handle specific forms of reasoning, such as common-sense or mathematical inference. As highlighted in RQ3, smaller models (2B--7B) are especially vulnerable to computational reasoning tasks like GSM8K, exhibiting sharper performance drops. Moreover, as noted in RQ2, different quantization methods can produce varying degrees of accuracy loss, making it difficult to draw definitive conclusions from a single task perspective.

\begin{simplebox}
\textbf{RQ4 Findings:} Difficult tasks do not always suffer the largest accuracy loss under quantization. The impact varies by model architecture and the chosen quantization method, leading some \textit{hard} tasks to remain stable while \textit{easier} tasks occasionally show bigger drops. In essence, quantization amplifies a model's existing weaknesses, particularly in commonsense, logical, or mathematical reasoning.
\end{simplebox}

\begin{figure}[ht]
	\centering
        \includegraphics[width=1\columnwidth]{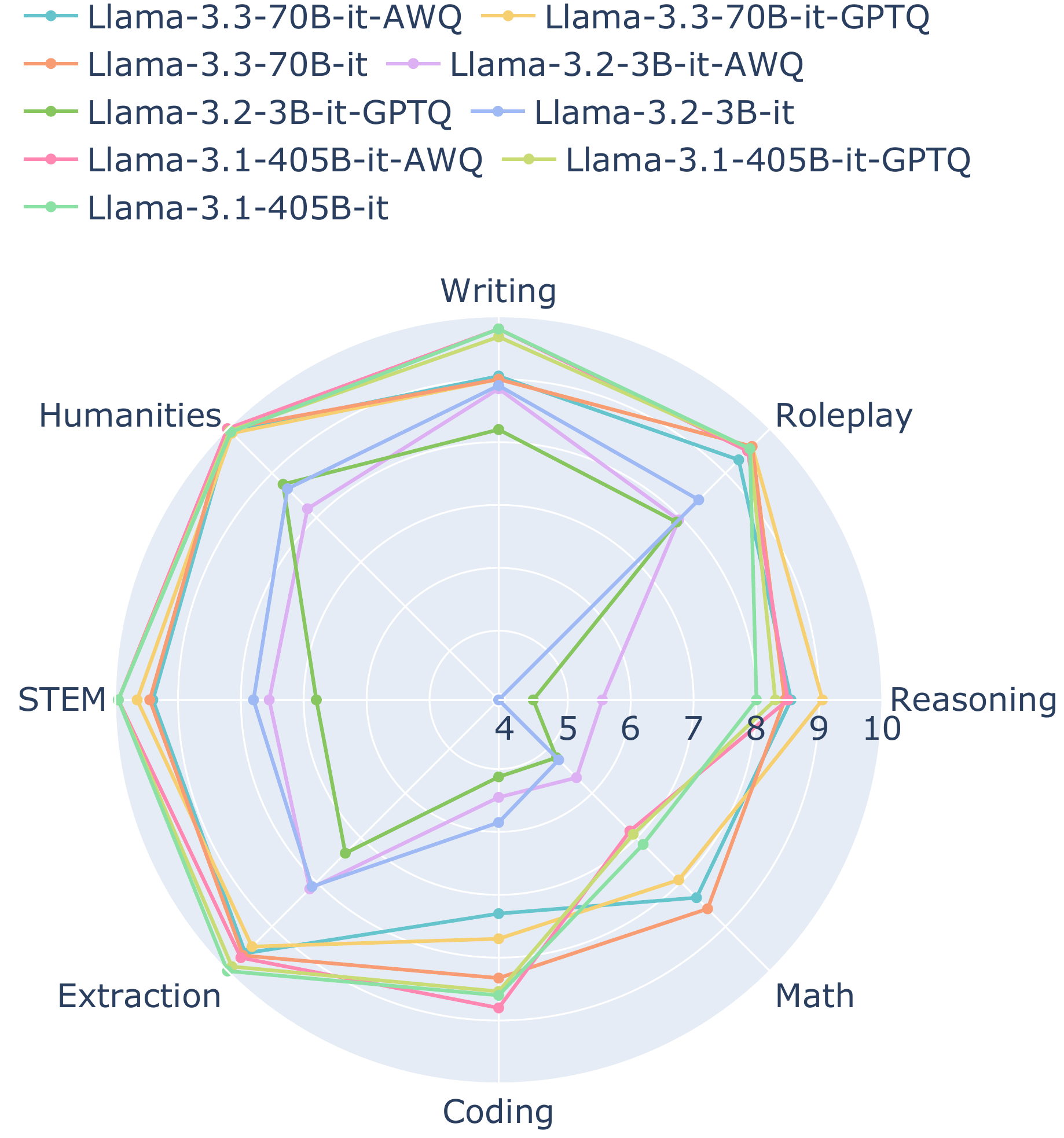}
        \caption{Category-wise MT-Bench scores of three quantized LLMs (Llama-3.3, Llama-3.2, and Llama-3.1) evaluated using AWQ and GPTQ methods. It highlights the performance differences across categories, including Writing, Roleplay, Reasoning, Math, Coding, Extraction, STEM, and Humanities, demonstrating the impact of quantization on diverse tasks.}
	\label{fig:rader_mt-bench}
\end{figure}

\subsection{RQ5: How does quantization impact the free-form conversation quality of LLMs when evaluated using the MT-Bench framework, which employs LLMs as judges?}

\noindent\textbf{Category-Level Analysis.}
Figure~\ref{fig:rader_mt-bench} presents a detailed breakdown of three models (Llama-3.1, Llama-3.2, and Llama-3.3) across MT-Bench categories. 
Quantized LLMs suffer the largest score degradation in \emph{Coding} and \emph{STEM}. 
For \textit{Coding}, manual inspections reveal that GPT4 often assigns lower scores when the generated code contains fewer examples or insufficient comments. 
In \textit{STEM} tasks, correctness matters when a definite solution exists, and concise logical explanations are important when an exact answer does not exist. 
Quantized models frequently either provided overly verbose justifications or produced incorrect statements, leading to lower scores. 
In contrast, the \emph{Reasoning} category showed an increase of about 1 point with quantization. Manual checks reveal that concise answers tend to receive higher GPT4 scores, and quantized models often responded more concisely than their original counterparts.
The detailed results of this manual inspection, along with the full text responses, can be found in Appendix~\ref{app:mt-qualitative_analysis}.
Also, Table~\ref{tab:mt-bench-categoray-full} in the Appendix~\ref{app:category_level_analysis} lists category-wise MT-Bench scores for all models.

\noindent\textbf{Limitations of GPT4-Based Evaluation.}
Consistent with the findings in MT-Bench, GPT4 sometimes misjudges incorrect responses as correct, particularly for math and reasoning tasks. 
Although reference-guided judging and chain-of-thought prompting can mitigate such errors~\cite{zheng2023judging}, they do not eliminate them entirely. 
In \textit{Reasoning} task, GPT4 erroneously considered a wrong answer correct, boosting quantized models' scores. 
Conversely, there were \textit{STEM} questions where both original and quantized models provided accurate answers, yet GPT4 mistakenly marked them as incorrect. 
\textbf{These misjudged cases are described in the Appendix~\ref{app:misjudged_cases}.}

For models like Llama-3.3-70B, categories such as \textit{Humanities} consistently scored near or at the maximum (10 points). 
In these cases, quantized versions also achieved scores in the high 9-point range, making it difficult to discern meaningful quality differences through manual inspection. 
This is because it is difficult to clearly understand the reasoning behind the results by only looking at GPT4’s judgment statements, especially when the score differences are marginal, such as 1-2 points.
Hence, for the latest large-scale models, more sensitive metrics or superior judging models may be required to evaluate subtle quality gaps.

\noindent\textbf{Multi-Turn Analysis.}
Table~\ref{tab:mt-bench} in the Appendix~\ref{app:model_level_analysis_and_multi_turn_results} presents the average scores of multi-turn conversations across 12 models. Among smaller models (e.g., 2B, 7B, 8B), some exhibit slight score improvements; however, this trend is not consistent across all small-scale models. In contrast, larger models (e.g., 13B, 70B) generally experience score declines, although there are exceptions where AWQ enhances performance. Additionally, accuracy losses become more noticeable in the second turn of multi-turn interactions.

These observations indicate that establishing a clear trend based solely on model size or type is challenging, as the impact of quantization depends on a combination of factors, including model architecture, quantization method, and task complexity. Furthermore, when comparing quantization methods, AWQ typically outperforms GPTQ, which is consistent with the results observed on OpenLLM Leaderboard-v1 and v2.

\noindent\textbf{Comparison with Leaderboard Results.}
The trends observed in MT-Bench do not always align with leaderboard outcomes. 
For instance, the Quantized Llama-3.1-405B model outperformed a newer Llama-3.3-70B-FP16 model on certain leaderboards, yet scored similarly or slightly lower in MT-Bench. 
Unlike the leaderboard tasks, which may not be particularly sensitive to \textit{Coding} and \textit{Math} challenges, the free-form and demanding nature of MT-Bench conversations highlights performance drops in more complex categories such as \textit{Coding} and \textit{STEM}. 
Thus, although computational metrics suggest that quantization does not uniformly degrade accuracy (RQ4), MT-Bench's qualitative LLM-based evaluation reveals significant performance reductions in tasks known to be difficult.

\begin{simplebox}
\textbf{RQ5 Findings}: We observe that quantization considerably reduces performance in Coding and STEM tasks, while occasionally improving reasoning.
The impact of quantization on multi-turn conversation quality does not consistently correlate with model size or type
Additionally, GPT4-based assessments sometimes misjudge incorrect answers as correct.
\end{simplebox}

\section{Conclusion}
We evaluated instruction-tuned quantized LLMs across 13 datasets and 6 task types, using models ranging from 1B to 405B and 4 quantization methods, including GPTQ, AWQ, SmoothQuant, and FP8.
We found that quantized LLMs generally outperformed smaller models in most tasks, except for hallucination detection and instruction-following. Performance varied by quantization method and precision, with weight-only quantization performing better in the 405B model. Task difficulty had little impact on accuracy loss.
Our MT-Bench evaluation revealed that quantization significantly reduces performance in Coding and STEM tasks while occasionally enhancing reasoning. Additionally, GPT4-based assessments can misjudge incorrect answers as correct.

\section*{Acknowledgments}
We thank the former Neubla ML team members — Minwook Ahn, Minho Park, Jinsol Kim, Raegeun Park, and Byonghwa Oh — for their valuable discussions and feedback.
This work was supported by the Institute of Information \& Communications Technology Planning \& Evaluation (IITP) grant funded by the Korea government (MSIT) (No.RS-2023-00277060, Development of open edge AI SoC hardware and software platform, 
No.RS-2024-00459797, Development of ML compiler framework for on-device AI, 
RS-2025-02214497, Development of low-level optimization program API technology for AI semiconductors)

\section*{Ethical Statement}
There are no ethical issues.

\bibliographystyle{named}
\bibliography{ijcai25}

\clearpage

\onecolumn
\appendix
\appendixpage
\addappheadtotoc

\begingroup
\renewcommand{\cftsecfont}{\bfseries} 
\renewcommand{\cftsecpagefont}{\bfseries} 
\setlength{\cftbeforesecskip}{0.5em} 
\renewcommand{\contentsname}{Table of Contents} 
\tableofcontents
\listoffigures      
\listoftables       
\endgroup

\captionsetup[figure]{list=yes}
\captionsetup[table]{list=yes}
\addtocontents{toc}{\protect\setcounter{tocdepth}{2}}

\break

\section{Details of Datasets}
\label{sec:details_datasets}
The 12 datasets used in the experiments are summarized in Table~\ref{tab:benchmarks_summary}.  Each dataset is described in detail below.

\noindent \textbf{Common-sense Reasoning.}
Among these, ARC, HellaSwag, and Winogrande target common-sense reasoning that is evident to humans yet remains challenging for AI. Notably, for ARC, we evaluated only the \textit{challenge} subset to focus on the most challenging questions.

\noindent \textbf{MMLU.}  
The MMLU dataset provides multiple-choice questions spanning 57 subjects, ranging from STEM fields to the humanities. Each question has four answer choices, and the dataset tests a model’s ability to handle a wide variety of topics with moderate to high difficulty.  

\noindent \textbf{GPQA.}  
GPQA consists of PhD-level queries formulated by domain experts. Since many questions require specialized knowledge, this dataset is particularly challenging for models lacking deep understanding in advanced subjects.

\noindent \textbf{MMLU-PRO.}  
MMLU-PRO is an extended variant of MMLU. It expands each question from four to ten answer choices and covers 14 topics, thereby increasing difficulty and necessitating more nuanced reasoning.

\noindent \textbf{BBH.}  
BBH comprises 23 subjects, including multi-step arithmetic, algorithmic reasoning, and satire. It evaluates a model’s capacity for human-like language understanding and logical processing.

\noindent \textbf{MuSR.}  
MuSR is divided into three tasks: murder mysteries, object placements, and team allocation. Solving these problems relies heavily on Chain-of-Thought (CoT) capabilities, as models must parse lengthy contexts, reason step by step, and combine partial conclusions.

\vspace{1em}
\noindent \textbf{IFEval.}  
IFEval tests whether models accurately follow instructions. It focuses purely on adherence to directives rather than content generation. For example, if a prompt requires writing over 400 words, the model should satisfy this requirement regardless of what it writes.

\noindent \textbf{TruthfulQA.}  
TruthfulQA evaluates how factually correct a model’s response is, serving as a benchmark for gauging hallucination tendencies. Models that produce incorrect or misleading outputs are penalized in this dataset.

\vspace{1em}
\noindent \textbf{GSM8K and MATH-Lvl-5.}  
GSM8K includes elementary-level math problems that generally need multi-step reasoning. MATH-Lvl-5 contains high-school competition problems across seven subjects, requiring LaTeX notation for equations and Asymptote for figures. These problems test a model’s ability to handle complex calculations and formatting.

\vspace{1em}
\noindent \textbf{MT-Bench.}  
MT-Bench employs GPT4 as a judge, using a single-answer grading system to evaluate multi-turn dialogues across eight categories (e.g., Writing, Roleplay, Reasoning). This free-form conversation benchmark emphasizes coherence, depth, and adaptability, making it well-suited for assessing real-world usage scenarios.

\vspace{1em}
\noindent \textbf{Alignment with OpenLLM Leaderboards.}  
We selected 12 benchmarks that align with the Huggingface OpenLLM Leaderboard-v1 (covering April 2023--June 2024) and Leaderboard-v2 (launched on June 26, 2024). This alignment helps reduce data contamination in newer models. For each task, we adopted the same few-shot In-Context Learning (ICL) settings used in the leaderboards~\cite{liu2023emergent,li2024evaluating}, ensuring consistent comparisons. Since Leaderboard-v2 does not yet include quantized model results and Leaderboard-v1 only features limited GPTQ examples, we performed all measurements in the same environment~\cite{LLMC2024} for a fair comparison.

\vspace{1em}
\noindent Overall, these datasets cover a wide spectrum of reasoning abilities, knowledge domains, and linguistic complexities, allowing us to rigorously evaluate how quantization impacts instruction-tuned LLMs across different tasks.

\begin{table*}[h]
    \centering
    \resizebox{\textwidth}{!}{%
        \begin{tabular}{
                >{\centering}m{0.3\textwidth} 
                >{\centering}m{0.4\textwidth} 
                >{\centering}m{0.1\textwidth} 
                >{\centering}m{0.15\textwidth}  
                >{\centering}m{0.5\textwidth} 
                >{\centering\arraybackslash}
                m{0.2\textwidth}
            }
            \toprule
            \textbf{Type} & \textbf{Dataset} & \textbf{\# Tasks} & \textbf{\# Test Samples} & \textbf{Evaluation Ability} & \textbf{Metrics} \\
            \midrule
            \multirow{3}{*}{CommonSenseQA} & ARC-Challenge~\cite{clark2018think} & 1 & 1,172 & Primary School Knowledge, ICL & Accuracy $\uparrow$ \\
            & HellaSwag~\cite{zellers2019hellaswag} & 1 &  10,003 & Knowledge, ICL & Accuracy $\uparrow$ \\
            & Winogrande~\cite{sakaguchi2021winogrande} & 1 & 1,767 & Knowledge, ICL & Accuracy $\uparrow$  \\
            \midrule
            \multirow{6}{*}{\parbox{0.35\textwidth}{\centering Complex Knowledge \& \newline Language Understanding}} & MMLU~\cite{hendryckstest2021} & 57 & 14,042 & Knowledge, ICL & Accuracy $\uparrow$ \\
            & GPQA~\cite{rein2023gpqa} & 3 &  448 & PhD-Lvl Knowledge & Accuracy $\uparrow$  \\
            & MMLU-PRO~\cite{wang2024mmlu} & 13 & 12,032  & Expert-Lvl Knowledge, ICL & Accuracy $\uparrow$ \\
            & BBH~\cite{suzgun2022challenging} & 23 &  6,511 & Knowledge, Algorithmic Reasoning, Language Understanding, ICL  & Accuracy $\uparrow$ \\
            & MuSR~\cite{sprague2023musr} & 3 & 756  & Reasoning, CoT & Accuracy $\uparrow$ \\
            \midrule
            \multirow{1}{*}{Instruction Following} & IFEval~\cite{ifeval} & 1 & 541 & Rigorousness, IF & Strict Accuracy $\uparrow$ \\
            \midrule
            \multirow{1}{*}{Hallucination} & TruthfulQA~\cite{lin2021truthfulqa} & 38 & 817 & Truthfulness & Multi-True (MC2) $\uparrow$ \\
            \midrule
            \multirow{2}{*}{\parbox{0.35\textwidth}{\centering Mathematics}} & GSM8K~\cite{cobbe2021gsm8k} & 1  & 1,319 & Mathematical Reasoning, CoT & Strict-Match $\uparrow$  \\
            & MATH-Lvl-5~\cite{hendrycks2021measuring,lewkowycz2022solving} & 7 & 12,000 & Mathematical Reasoning, CoT & Exact-Match $\uparrow$ \\
            \midrule              
            \multirow{1}{*}{Dialogue} & MT-Bench~\cite{zheng2023judging} & 8 & 80 & Conversation Quality & Score (0--10) $\uparrow$\\
            \bottomrule
        \end{tabular}
    }
    \caption{Overview of different datasets, their type, \# of tasks, \# of samples, evaluation ability, and assessment metrics. ICL, CoT, and IF are acronyms for In-Context Learning, Chain-of-Thought, and Instruction Following, respectively, which are emergent abilities.}
    \label{tab:benchmarks_summary}
\end{table*}

\section{Additional Models}
\label{app:additional_models}
Table~\ref{tab:evaluation_results_other_models} and Table~\ref{tab:evaluation_results_v2_other_models} demonstrate how GPTQ and AWQ quantization methods affect the accuracy of Vicuna and Gemma across OpenLLM Leaderboard-v1 and OpenLLM Leaderboard-v2. 
Overall, AWQ tends to incur smaller accuracy drops than GPTQ when compared to the FP16 baseline, particularly on tasks such as ARC-c and GSM8k. In contrast, GPTQ often exhibits more noticeable performance declines. 
This pattern persists in both Leaderboard-v1 (covering ARC-c, GSM8k, HellaSwag, MMLU, TruthfulQA, Winogrande) and Leaderboard-v2 (covering BBH, GPQA, IFEval, Math-Lvl-5, MMLU-PRO, MuSR), suggesting that AWQ maintains more stable results across a variety of tasks.

\begin{table*}[ht]
    \centering
    \begin{threeparttable}[t]
    \resizebox{\textwidth}{!}{
        \begin{tabular}{cccccccccccc}
            \toprule
            \multirow{4.5}{*}{\textbf{Model}} & \multirow{4.5}{*}{\textbf{Method}} & \multirow{4.5}{*}{\textbf{W/A}} & \multirow{4.5}{*}{\makecell{\textbf{Storage} \\ (GB)}} 
                & \multicolumn{6}{c}{\textbf{OpenLLM Leaderboard-v1 $\uparrow$}} \\
            \cmidrule(lr){5-11}
                & & & & \headerthree{ARC-c}{25-shot}{acc\_norm} & \headerthree{GSM8k}{5-shot}{acc} & \headerthree{HellaSwag}{10-shot}{acc\_norm} & \headerthree{MMLU}{5-shot}{acc} & \headerthree{TruthfulQA}{0-shot}{mc2} & \headerthree{Winogrande}{5-shot}{acc} & \textbf{Avg.} \\
            \midrule
            \multirow{3}{*}{Vicuna-7B-v1.3} 
                & FP16 & 16 / 16 & 14 & 51.28 & 13.34 & 77.36 & 47.21 & 47.01 & 71.03 & 51.21 \\
                & GPTQ\(^*\) & 4 / 16 & 3.5 & \vone{46.84}{-4.44} & \vone{10.31}{-3.03} & \vone{75.06}{-2.30} & \vone{43.54}{-3.67} & \vone{44.18}{-2.83} & \vone{69.06}{-1.97} & \vone{48.17}{-3.04} \\
                & AWQ & 4 / 16 & 3.5 & \vone{51.20}{-0.08} & \vone{12.74}{-0.60} & \vone{76.63}{-0.73} & \vone{46.34}{-0.87} & \vone{46.65}{-0.36} & \vone{71.35}{0.32} & \vone{50.82}{-0.39} \\
            \midrule

            \multirow{3}{*}{Gemma-2B-it} 
                & FP16 & 16 / 16 & 4 & 44.03 & 4.01 & 63.01 & 36.92 & 45.76 & 61.48 & 42.54 \\
                & GPTQ\(^*\) & 4 / 16 & 1 & \vone{44.28}{0.25} & \vone{3.10}{-0.91} & \vone{61.76}{-1.25} & \vone{37.21}{0.29} & \vone{46.31}{0.55} & \vone{60.14}{-1.34} & \vone{42.13}{-0.41} \\
                & AWQ & 4 / 16 & 1 & \vone{42.74}{-1.29} & \vone{1.66}{-2.35} & \vone{60.98}{-2.03} & \vone{36.21}{-0.71} & \vone{45.18}{-0.58} & \vone{59.66}{-1.82} & \vone{41.07}{-1.47} \\
            \midrule

            \multirow{3}{*}{Gemma-7B-it}
                & FP16 & 16 / 16 & 14 & 51.54 & 27.45 & 72.06 & 51.62 & 47.05 & 64.72 & 52.41 \\
                & GPTQ\(^*\) & 4 / 16 & 3.5 & \vone{51.02}{-0.52} & \vone{25.62}{-1.83} & \vone{70.75}{-1.31} & \vone{51.21}{-0.41} & \vone{47.11}{0.06} & \vone{66.29}{1.57} & \vone{52.00}{-0.41} \\
                & AWQ & 4 / 16 & 3.5 & \vone{50.43}{-1.11} & \vone{25.85}{-1.60} & \vone{71.27}{-0.79} & \vone{51.14}{-0.48} & \vone{46.11}{-0.94} & \vone{66.37}{1.65} & \vone{51.86}{-0.55} \\   
            \midrule

        \end{tabular}
    }
    \end{threeparttable}%
    \caption{Evaluation of Vicuna and Gemma on OpenLLM Leaderboard-v1 tasks including ARC-c, GSM8k, HellaSwag, MMLU, TruthfulQA, and Winogrande. \(^*\) denotes the use of AutoGPTQ for quantization.}
    \label{tab:evaluation_results_other_models}
\end{table*}

\begin{table*}[ht]
    \centering
    \begin{threeparttable}[t]
    \resizebox{\textwidth}{!}{
        \begin{tabular}{cccccccccccc}
            \toprule
            \multirow{3.5}{*}{\textbf{Model}} & \multirow{3.5}{*}{\textbf{Method}} & \multirow{3.5}{*}{\textbf{W/A}} & \multirow{3.5}{*}{\makecell{\textbf{Storage} \\ (GB)}} 
                & \multicolumn{6}{c}{\textbf{OpenLLM Leaderboard-v2 $\uparrow$}} \\
            \cmidrule(lr){5-11}
                & & & & \headertwo{BBH}{3-shot} & \headertwo{GPQA}{0-shot} & \headertwo{IFEval}{0-shot} & \headertwo{Math-Lvl-5}{4-shot} & \headertwo{MMLU-PRO}{5-shot} & \headertwo{MuSR}{0-shot} & \textbf{Avg.} \\
            \midrule
            \multirow{3}{*}{Vicuna-7B-v1.3} 
                & FP16 & 16 / 16 & 14 & 12.26 & 2.08 & 26.13 & 1.61 & 10.03 & 5.72 & 9.64 \\
                & GPTQ\(^*\) & 4 / 16 & 3.5 & \vtwo{7.61}{-4.65} & \vtwo{0.12}{-1.96} & \vtwo{28.07}{1.94} & \vtwo{0.79}{-0.82} & \vtwo{6.40}{-3.63} & \vtwo{3.85}{-1.87} & \vtwo{7.81}{-1.83} \\
                & AWQ & 4 / 16 & 3.5 & \vtwo{10.65}{-1.61} & \vtwo{0.61}{-1.47} & \vtwo{22.90}{-3.23} & \vtwo{0.82}{-0.79} & \vtwo{8.52}{-1.51} & \vtwo{3.15}{-2.57} & \vtwo{7.78}{-1.86} \\
            \midrule

           \multirow{3}{*}{Gemma-2B-it} 
                & FP16 & 16 / 16 & 4 & 4.67 & 6.72 & 27.02 & 2.01 & 6.23 & 4.82 & 8.58 \\
                & GPTQ\(^*\) & 4 / 16 & 1 & \vtwo{4.16}{-0.51} & \vtwo{6.72}{0.00} & \vtwo{28.76}{1.74} & \vtwo{1.38}{-0.63} & \vtwo{5.94}{-0.29} & \vtwo{4.81}{-0.01} & \vtwo{8.63}{-0.05} \\
                & AWQ & 4 / 16 & 1 & \vtwo{4.96}{0.29} & \vtwo{3.79}{-2.93} & \vtwo{29.02}{2.00} & \vtwo{1.39}{-0.62} & \vtwo{5.69}{-0.54} & \vtwo{3.92}{-0.90} & \vtwo{8.13}{-0.45} \\
            \midrule

            \multirow{3}{*}{Gemma-7B-it}
                & FP16 & 16 / 16 & 14 & 15.89 & 3.54 & 35.68 & 2.65 & 14.55 & 17.00 & 14.89 \\
                & GPTQ\(^*\) & 4 / 16 & 3.5 & \vtwo{13.51}{-2.38} & \vtwo{9.65}{6.11} & \vtwo{35.56}{-0.12} & \vtwo{1.92}{-0.73} & \vtwo{13.66}{-0.89} & \vtwo{18.26}{1.26} & \vtwo{15.43}{0.54} \\
                & AWQ & 4 / 16 & 3.5 & \vtwo{15.63}{-0.26} & \vtwo{6.23}{2.69} & \vtwo{36.39}{0.71} & \vtwo{2.40}{-0.25} & \vtwo{13.76}{-0.79} & \vtwo{15.73}{-1.27} & \vtwo{15.02}{0.13} \\
            \midrule
        \end{tabular}
    }
    \end{threeparttable}%
    \caption{Evaluation of Vicuna and Gemma on OpenLLM Leaderboard-v2 tasks including BBH, GPQA, IFEval, Math-Lvl-5, MMLU-PRO, and MuSR. \(^*\) denotes the use of AutoGPTQ for quantization.}
    \label{tab:evaluation_results_v2_other_models}
\end{table*}

\definecolor{upcolor}{HTML}{DDFFDD}   
\definecolor{downcolor}{HTML}{FFD9D9} 

\begin{table*}[ht]
\centering
\resizebox{\textwidth}{!}{
\begin{tabular}{llccccccccc}
\toprule
\textbf{Model} & \textbf{Method}
 & \textbf{Writing} 
 & \textbf{Roleplay} 
 & \textbf{Reasoning} 
 & \textbf{Math} 
 & \textbf{Coding} 
 & \textbf{Extraction} 
 & \textbf{STEM} 
 & \textbf{Humanities} 
 & \textbf{Avg.} \\
\midrule

\multirow{3}{*}{Vicuna-7B-v1.3}
 & \textbf{FP16}
   & 8.10 & 7.45 & 4.65 & 2.30 & 3.55 & 5.00 & 7.83 & 9.10 & 6.00 \\
 & GPTQ 
   & -0.05 
   & -0.50 
   & \cellcolor{upcolor}{+0.20}
   & -0.40 
   & \cellcolor{downcolor}{-0.95}
   & \cellcolor{upcolor}{+0.45}
   & \cellcolor{downcolor}{-0.53}
   & \cellcolor{downcolor}{-0.55}
   & -0.29 \\
 & AWQ 
   & \cellcolor{upcolor}{+0.20}
   & \cellcolor{upcolor}{+0.30}
   & \cellcolor{upcolor}{+0.70}
   & \cellcolor{upcolor}{+0.75}
   & \cellcolor{downcolor}{-1.00}
   & \cellcolor{downcolor}{-0.55}
   & \cellcolor{upcolor}{+0.07}
   & \cellcolor{downcolor}{-0.20}
   & +0.03 \\ 
\midrule

\multirow{3}{*}{Gemma-2B-it}
 & \textbf{FP16}
   & 6.67 & 5.90 & 2.75 & 1.75 & 2.40 & 4.45 & 6.05 & 7.10 & 4.63 \\
 & GPTQ
   & \cellcolor{downcolor}{-0.42}
   & -0.20
   & \cellcolor{upcolor}{+0.30}
   & \cellcolor{upcolor}{+0.55}
   & \cellcolor{downcolor}{-0.65}
   & \cellcolor{upcolor}{+0.15}
   & -0.25
   & \cellcolor{downcolor}{-0.32}
   & -0.10 \\
 & AWQ
   & \cellcolor{downcolor}{-0.70}
   & \cellcolor{upcolor}{+0.45}
   & -0.05
   & \cellcolor{upcolor}{+0.05}
   & \cellcolor{downcolor}{-0.75}
   & \cellcolor{downcolor}{-0.65}
   & \cellcolor{upcolor}{+0.30}
   & -0.25
   & -0.20 \\
\midrule

\multirow{3}{*}{Gemma-7B-it}
 & \textbf{FP16}
   & 6.97 & 6.60 & 5.00 & 3.70 & 3.75 & 5.90 & 7.05 & 8.93 & 5.99 \\
 & GPTQ
   & \cellcolor{upcolor}{+0.11}
   & -0.60
   & -0.15
   & \cellcolor{downcolor}{-1.10}
   & \cellcolor{downcolor}{-0.70}
   & \cellcolor{upcolor}{+0.35}
   & -0.13
   & \cellcolor{downcolor}{-1.18}
   & -0.43 \\
 & AWQ
   & -0.09
   & \cellcolor{downcolor}{-0.70}
   & -0.35
   & \cellcolor{downcolor}{-0.90}
   & \cellcolor{downcolor}{-0.40}
   & -0.40
   & \cellcolor{upcolor}{+0.05}
   & -0.03
   & -0.36 \\
\midrule

\multirow{3}{*}{Llama-2-7B-Chat}
 & \textbf{FP16}
   & 8.43 & 7.85 & 5.00 & 2.50 & 3.10 & 6.62 & 8.97 & 9.85 & 6.54 \\
 & GPTQ
   & -0.43
   & -0.55
   & -0.50
   & \cellcolor{upcolor}{+0.10}
   & -0.65
   & \cellcolor{downcolor}{-1.27}
   & \cellcolor{downcolor}{-0.82}
   & \cellcolor{downcolor}{-1.42}
   & -0.69 \\
 & AWQ
   & \cellcolor{upcolor}{+0.87}
   & \cellcolor{downcolor}-0.45
   & \cellcolor{upcolor}{+0.05}
   & \cellcolor{upcolor}{+0.30}
   & \cellcolor{downcolor}{-0.60}
   & \cellcolor{upcolor}{+0.63}
   & -0.02
   & \cellcolor{downcolor}-0.15
   & +0.08 \\
\midrule

\multirow{3}{*}{Llama-2-13B-Chat}
 & \textbf{FP16}
   & 8.90 & 7.80 & 6.15 & 2.50 & 4.20 & 8.10 & 8.88 & 9.85 & 7.05 \\
 & GPTQ
   & \cellcolor{upcolor}{+0.50}
   & -0.50
   & \cellcolor{downcolor}{-0.90}
   & \cellcolor{upcolor}{+0.65}
   & \cellcolor{downcolor}-0.75
   & \cellcolor{downcolor}{-1.00}
   & \cellcolor{upcolor}{+0.30}
   & 0.00
   & -0.22 \\
 & AWQ
   & -0.12
   & \cellcolor{downcolor}-0.25
   & 0.00
   & \cellcolor{upcolor}{+0.50}
   & \cellcolor{downcolor}-0.60
   & \cellcolor{downcolor}{-0.75}
   & \cellcolor{upcolor}{+0.05}
   & \cellcolor{upcolor}{+0.05}
   & -0.14 \\
\midrule

\multirow{3}{*}{Llama-2-70B-Chat}
 & \textbf{FP16}
   & 9.25 & 7.80 & 6.35 & 4.05 & 3.74 & 7.40 & 8.95 & 9.90 & 7.18 \\
 & GPTQ
   & \cellcolor{upcolor}{+0.05}
   & \cellcolor{upcolor}{+0.03}
   & \cellcolor{upcolor}{+0.10}
   & \cellcolor{downcolor}-0.80
   & \cellcolor{upcolor}{+0.06}
   & 0.00
   & \cellcolor{downcolor}-0.25
   & 0.00
   & -0.10 \\
 & AWQ
   & -0.18
   & \cellcolor{upcolor}{+0.80}
   & \cellcolor{downcolor}{-0.95}
   & \cellcolor{upcolor}{+0.15}
   & \cellcolor{downcolor}-0.34
   & \cellcolor{upcolor}{+0.20}
   & \cellcolor{downcolor}-0.40
   & \cellcolor{downcolor}-0.28
   & -0.13 \\
\midrule

\multirow{3}{*}{Llama-3.1-8B-it}
 & \textbf{FP16}
   & 9.35 & 9.15 & 4.60 & 6.60 & 6.40 & 8.62 & 8.45 & 9.95 & 7.89 \\
 & GPTQ
   & \cellcolor{downcolor}-0.25
   & \cellcolor{downcolor}{-1.25}
   & -0.05
   & \cellcolor{upcolor}{+0.50}
   & \cellcolor{downcolor}{-1.15}
   & \cellcolor{upcolor}{+0.03}
   & \cellcolor{upcolor}{+0.73}
   & \cellcolor{downcolor}-0.55
   & -0.25 \\
 & AWQ
   & \cellcolor{downcolor}-0.60
   & \cellcolor{upcolor}{+0.10}
   & \cellcolor{upcolor}{+1.45}
   & \cellcolor{downcolor}-0.95
   & \cellcolor{upcolor}{+0.05}
   & \cellcolor{downcolor}-0.82
   & \cellcolor{upcolor}{+1.17}
   & 0.00
   & +0.05 \\
\midrule

\multirow{3}{*}{Llama-3.1-70B-it}
 & \textbf{FP16}
   & 9.65 & 9.50 & 7.58 & 7.65 & 8.14 & 9.70 & 9.93 & 9.99 & 9.02 \\
 & GPTQ
   & \cellcolor{downcolor}-0.15
   & \cellcolor{downcolor}{-0.70}
   & \cellcolor{upcolor}{+0.22}
   & \cellcolor{downcolor}-0.35
   & \cellcolor{upcolor}{+0.16}
   & \cellcolor{upcolor}{+0.05}
   & -0.13
   & -0.02
   & -0.12 \\
 & AWQ
   & \cellcolor{upcolor}{+0.10}
   & 0.00
   & \cellcolor{upcolor}{+0.67}
   & \cellcolor{upcolor}{+0.10}
   & \cellcolor{downcolor}-0.19
   & \cellcolor{downcolor}-0.20
   & \cellcolor{upcolor}{+0.02}
   & \cellcolor{downcolor}-0.04
   & +0.06 \\
\midrule


\multirow{3}{*}{Llama-3.1-405B-it}
 & \textbf{FP16}
   & 9.80 & 9.55 & 8.00 & 7.15 & 8.60 & 10.00 & 9.95 & 9.93 & 9.12 \\
 & GPTQ
   & \cellcolor{downcolor}{-0.12} & 0.00 &  \cellcolor{upcolor}{+0.30} &  \cellcolor{downcolor}{-0.23} &  -0.06 &  \cellcolor{downcolor}{-0.10} &  0.00 &  \cellcolor{upcolor}{+0.02} &  -0.02 \\
& AWQ
    & 0.00 & \cellcolor{downcolor}{-0.05} & \cellcolor{upcolor}{+0.50} & \cellcolor{downcolor}{-0.30} & \cellcolor{upcolor}{+0.20} & \cellcolor{downcolor}{-0.30} & 0.00 & \cellcolor{upcolor}{+0.07} & +0.02 \\
\midrule

\multirow{3}{*}{Llama-3.2-1B-it}
 & \textbf{FP16}
   & 7.05 & 6.35 & 2.40 & 3.80 & 3.40 & 4.15 & 6.15 & 8.50 & 5.22 \\
 & GPTQ
   & \cellcolor{downcolor}{-1.30}
   & -0.30
   & \cellcolor{upcolor}{+1.00}
   & \cellcolor{downcolor}{-1.10}
   & -0.24
   & \cellcolor{upcolor}{+1.10}
   & -0.50
   & \cellcolor{downcolor}{-1.45}
   & -0.34 \\
 & AWQ
   & \cellcolor{downcolor}-0.35
   & \cellcolor{upcolor}{+0.70}
   & \cellcolor{upcolor}{+0.95}
   & -0.32
   & -0.25
   & \cellcolor{downcolor}-0.60
   & -0.07
   & \cellcolor{downcolor}-0.50
   & -0.05 \\
\midrule

\multirow{3}{*}{Llama-3.2-3B-it}
 & \textbf{FP16}
   & 8.90 & 8.40 & 3.90 & 5.25 & 5.85 & 8.10 & 7.80 & 8.65 & 7.11 \\
 & GPTQ
   & -0.70
   & -0.50
   & \cellcolor{upcolor}{+0.55}
   & -0.05
   & \cellcolor{downcolor}-0.73
   & \cellcolor{downcolor}-0.75
   & \cellcolor{downcolor}{-1.00}
   & \cellcolor{upcolor}{+0.10}
   & -0.39 \\
 & AWQ
   & -0.05
   & \cellcolor{downcolor}-0.45
   & \cellcolor{upcolor}{+1.65}
   & \cellcolor{upcolor}{+0.40}
   & \cellcolor{downcolor}-0.40
   & \cellcolor{upcolor}{+0.05}
   & -0.25
   & \cellcolor{downcolor}-0.45
   & +0.06 \\
\midrule

\multirow{3}{*}{Llama-3.3-70B-it}
 & \textbf{FP16}
   & 9.00 & 9.60 & 8.45 & 8.60 & 8.32 & 9.65 & 9.45 & 10.00 & 9.13 \\
 & GPTQ
   & 0.00
   & 0.00
   & \cellcolor{upcolor}{+0.60}
   & \cellcolor{downcolor}-0.65
   & \cellcolor{downcolor}-0.62
   & \cellcolor{downcolor}-0.20
   & \cellcolor{upcolor}{+0.20}
   & -0.10
   & -0.09 \\
 & AWQ
   & \cellcolor{upcolor}{+0.05}
   & \cellcolor{downcolor}-0.30
   & \cellcolor{upcolor}{+0.10}
   & \cellcolor{downcolor}-0.25
   & \cellcolor{downcolor}{-1.02}
   & -0.05
   & -0.05
   & -0.05
   & -0.19 \\
\bottomrule
\end{tabular}
}
\caption{
For each category, the MT-Bench scores (FP16, GPTQ, AWQ) are presented for 12 models. The difference in MT-Bench scores compared to FP16 is computed and shown for GPTQ and AWQ. The three categories with the largest score drops compared to FP16 are highlighted in \colorbox[HTML]{FFD9D9}{red}, while The categories with score increases are highlighted in \colorbox[HTML]{DDFFDD}{green}.
}
\label{tab:mt-bench-categoray-full}
\end{table*}

\begin{table}[h]
    \centering
    \begin{tabular}{cccccc}
        \toprule
        \textbf{Model} & \textbf{Turn} & \textbf{FP16} & \makecell{\textbf{GPTQ}\(^*\) \\ (W4/A16)}  & \makecell{\textbf{AWQ} \\ (W4/A16)} & \makecell{\textbf{API} \\ \textbf{Cost}}  \\
        \midrule
        \multirow{3}{*}{Vicuna-7B-v1.3} 
            & 1 & \cellcolor[HTML]{CAFFCA}\textbf{6.69} & 6.11 & 6.56 & \multirow{3}{*}{\$4.96} \\
            & 2 & 5.30 & 5.30 & \cellcolor[HTML]{CAFFCA}\textbf{5.50} & \\
            & Avg. & 6.00 & 5.71 & \cellcolor[HTML]{CAFFCA}\textbf{6.03} & \\
        \midrule
        \multirow{3}{*}{Gemma-2B-it} 
            & 1 & \cellcolor[HTML]{CAFFCA}\textbf{5.09} & 5.03 & 4.68 & \multirow{3}{*}{\$4.47} \\
            & 2 & 4.18 & 4.03 & \cellcolor[HTML]{CAFFCA}\textbf{4.19} & \\
            & Avg. & \cellcolor[HTML]{CAFFCA}\textbf{4.63} & 4.53 & 4.43 & \\
        \midrule
        \multirow{3}{*}{Gemma-7B-it} 
            & 1 & \cellcolor[HTML]{CAFFCA}\textbf{6.89} & 6.40 & 6.34 & \multirow{3}{*}{\$4.47} \\
            & 2 & \cellcolor[HTML]{CAFFCA}\textbf{5.09} & 4.73 & 4.93 & \\
            & Avg. & \cellcolor[HTML]{CAFFCA}\textbf{5.99} & 5.56 & 5.63 & \\
        \midrule
        \multirow{3}{*}{Llama-2-7B-Chat} 
            & 1 & \cellcolor[HTML]{CAFFCA}\textbf{7.12} & 6.67 & 6.86 & \multirow{3}{*}{\$5.80} \\
            & 2 & 5.96 & 4.99 & \cellcolor[HTML]{CAFFCA}\textbf{6.38} & \\
            & Avg. & 6.54 & 5.85 & \cellcolor[HTML]{CAFFCA}\textbf{6.62} & \\
        \midrule
        \multirow{3}{*}{Llama-2-13B-Chat} 
            & 1 & 7.34 & 7.29 & \cellcolor[HTML]{CAFFCA}\textbf{7.38} & \multirow{3}{*}{\$5.85} \\
            & 2 & \cellcolor[HTML]{CAFFCA}\textbf{6.75} & 6.38 & 6.44 & \\
            & Avg. & \cellcolor[HTML]{CAFFCA}\textbf{7.05} & 6.83 & 6.91 & \\
        \midrule
        \multirow{3}{*}{Llama-2-70B-Chat} 
            & 1 & 7.28 & 7.21 & \cellcolor[HTML]{CAFFCA}\textbf{7.29} & \multirow{3}{*}{\$7.47} \\
            & 2 & \cellcolor[HTML]{CAFFCA}\textbf{7.13} & 6.95 & 6.83 & \\
            & Avg. & \cellcolor[HTML]{CAFFCA}\textbf{7.18} & 7.08 & 7.05 & \\
        \midrule
        \multirow{3}{*}{Llama-3.1-8B-it} 
            & 1 & 8.17 & 7.89 & \cellcolor[HTML]{CAFFCA}\textbf{8.52} & \multirow{3}{*}{\$5.49} \\ 
            & 2 & \cellcolor[HTML]{CAFFCA}\textbf{7.61} & 7.39 & 7.36 & \\
            & Avg. & 7.89 & 7.64 & \cellcolor[HTML]{CAFFCA}\textbf{7.94} & \\
        \midrule
        \multirow{3}{*}{Llama-3.1-70B-it} 
            & 1 & 9.24 & 9.04 & \cellcolor[HTML]{CAFFCA}\textbf{9.25} & \multirow{3}{*}{\$5.87} \\
            & 2 & 8.80 & 8.76 & \cellcolor[HTML]{CAFFCA}\textbf{8.90} & \\
            & Avg. & 9.02 & 8.90 & \cellcolor[HTML]{CAFFCA}\textbf{9.08} & \\
        \midrule
        \multirow{3}{*}{Llama-3.1-405B-it} 
            & 1 & \cellcolor[HTML]{CAFFCA}\textbf{9.29} & 9.28 & 9.24 & \multirow{3}{*}{\$5.56} \\
            & 2 & 8.95 & 8.88 & \cellcolor[HTML]{CAFFCA}\textbf{9.04} & \\
            & Avg. & 9.12 & 9.10 & \cellcolor[HTML]{CAFFCA}\textbf{9.14} & \\        
        \midrule
        \multirow{3}{*}{Llama-3.2-1B-it} 
            & 1 & \cellcolor[HTML]{CAFFCA}\textbf{5.71} & 5.50 & 5.56 & \multirow{3}{*}{\$5.06}
            \\
            & 2 & 4.74 & 4.27 & \cellcolor[HTML]{CAFFCA}\textbf{4.78} & \\
            & Avg. & \cellcolor[HTML]{CAFFCA}\textbf{5.22} & 4.88 & 5.17 & \\
        \midrule
        \multirow{3}{*}{Llama-3.2-3B-it} 
            & 1 & \cellcolor[HTML]{CAFFCA}\textbf{7.58} & 7.29 & 7.46 & \multirow{3}{*}{\$4.99}
            \\
            & 2 & 6.64 & 6.15 & \cellcolor[HTML]{CAFFCA}\textbf{6.88} & \\
            & Avg. & 7.11 & 6.72 & \cellcolor[HTML]{CAFFCA}\textbf{7.17} & \\
        \midrule
        \multirow{3}{*}{Llama-3.3-70B-it} 
            & 1 & \cellcolor[HTML]{CAFFCA}\textbf{9.14} & 9.13 & 9.01 & \multirow{3}{*}{\$5.70}
            \\
            & 2 & \cellcolor[HTML]{CAFFCA}\textbf{9.13} & 8.95 & 8.86 & \\
            & Avg. & \cellcolor[HTML]{CAFFCA}\textbf{9.13} & 9.04 & 8.94 & \\
        \bottomrule
    \end{tabular}
    \caption{A evaluation of the influence of quantization on the MT-Bench multi-turn conversation benchmark. \(^*\) denotes the use of AutoGPTQ for quantization. The API cost represents the GPT4 usage fee incurred when running MT-Bench for a single model.
    We highlighted the \colorbox[HTML]{CAFFCA}{best value} in green.}
    \label{tab:mt-bench}
\end{table}

\section{Detailed Results of MT-Bench}

\subsection{Category-Level Analysis}
\label{app:category_level_analysis}
Table~\ref{tab:mt-bench-categoray-full} shows that the Reasoning category improves for 9 out of 12 models after quantization. By contrast, the categories with the largest accuracy drops are Coding, Humanities, and Extraction. Among the 24 total quantized configurations (Models $\times$ Quantization), Coding, Humanities, and Extraction each appear 16, 12, and 10 times, respectively, among the top three categories with the most severe performance losses. When analyzing all eight MT-Bench categories under quantization, the most pronounced declines are observed in Math, Coding, and Reasoning. Llama-3.3-70B, the newest model, achieves notably higher scores across all eight categories.

\subsection{Model-Level Analysis and Multi-Turn Results}
\label{app:model_level_analysis_and_multi_turn_results}
Table~\ref{tab:mt-bench} presents the multi-turn performance for each model. While smaller models such as Llama-3.1-8B-it, Llama-3.2-1B, and Llama-3.2-3B sometimes show unexpected boosts in Reasoning (e.g., +1.65 in Llama-3.2-3B), larger models generally do not experience such gains. Although Llama-3.2 is among the latest releases, it does not exhibit markedly different multi-turn behavior compared to other recent models. Additionally, models like Llama-3.1 tend to rank highly in MT-Bench, possibly because of their similarity to GPT4 in certain tasks; however, these close scores make it difficult to treat the results as definitive. Quantized variants also perform well under this evaluation scheme.

\subsection{Cost for MT-Bench}
Across all models, the MT-Bench evaluation incurs a cost ranging from \$4 to \$7 each, totaling about \$200 in GPT4 API usage. This highlights the resource-intensive nature of multi-turn LLM evaluations, particularly when using state-of-the-art judging models.

\subsection{Qualitative Analysis}
\label{app:mt-qualitative_analysis}

\begin{complexbox}[label=reasoning102]{An improved example and a degraded example of the Reasoning question on MT-Bench by the quantized Llama-3.2-3B-it.}
\textbf{QUESTION:}

\textbf{Turn 1:} You can see a beautiful red house to your left and a hypnotic greenhouse to your right, an attractive heated pink place in the front. So, where is the White House?

\textbf{Turn 2:} Does the original question contain any clues to definitively determine the location of the White House?

\textbf{reference:}

\textbf{Turn 1:} \textcolor{blue}{The answer is Washington, DC.}

\textbf{Turn 2:} \textcolor{blue}{No.}

\noindent\rule{\linewidth}{0.5pt}

\textbf{FP16 (original) ANSWER:}  

\textbf{Turn 1 \textcolor{blue}{(correct)}:} The White House is actually located in \textcolor{blue}{Washington D.C.}, which is the capital city of the United States. [\ldots]
The red house you are referring to is likely the White House itself, \ldots




$\rightarrow$ \textbf{GPT4 Judgement: (6/10)} \ldots
If the references to the red house, greenhouse, and heated pink place are part of a riddle or a creative description, \textcolor{red}{additional context would be helpful} to provide a more accurate interpretation.

\textbf{Turn 2 \textcolor{blue}{(generally correct)}:} 


\ldots So, \textcolor{blue}{while the original question does not contain any explicit clues}, I think it is possible that the beautiful red house, hypnotic greenhouse, \ldots

$\rightarrow$ \textbf{GPT4 Judgement: (2/10)} \ldots 
\textcolor{red}{Descriptions such as a beautiful red house, a hypnotic greenhouse, and an attractive heated pink place do not relate to its actual location.} \ldots

\noindent\rule{\linewidth}{0.5pt}

\textbf{W4A16-GPTQ ANSWER (Degraded):}

\textbf{Turn 1 \textcolor{blue}{(correct)}:}
\textcolor{blue}{The White House is actually located in Washington, D.C.}, 
[\ldots] \textcolor{red}{\{But if you're looking for a building that is similar in style and architecture to the White House, you might be thinking of the U.S. Capitol Building's neighbor, the Jefferson Memorial's neighbor, the Arlington National Cemetery's neighbor... \} $ \times $ 11 }(This part is repeated 11 times).

$\rightarrow$ \textbf{GPT4 Judgement: (1/10)} \ldots
\textcolor{red}{the assistant provided a long, repetitive, and confusing response that did not directly answer the question.} \ldots

\textbf{Turn 2 \textcolor{red}{(wrong)}:} \ldots
\textcolor{red}{these hints suggest the White House is in Washington, D.C}.
While these interpretations are educated guesses, they align with the clues provided. 

$\rightarrow$ \textbf{GPT4 Judgement: (1/10)} \ldots
The assistant incorrectly interprets the descriptions of a "beautiful red house", a "hypnotic greenhouse", and an "attractive heated pink place" as clues to the location of the White House. \textcolor{red}{These descriptions are unrelated to the White House's location.}


\noindent\rule{\linewidth}{0.5pt}

\textbf{W4A16-AWQ ANSWER (Improved):}  

\textbf{Turn 1 \textcolor{blue}{(correct}:} \textcolor{blue}{The White House is actually located in Washington, D.C.}, \ldots

$\rightarrow$ \textbf{GPT4 Judgement: (10/10)} \ldots
\textcolor{blue}{It provides the same information as the reference answer}, \ldots

\textbf{Turn 2 \textcolor{blue}{(correct)}:} \ldots
\textcolor{blue}{these descriptions don't provide any specific clues to determine the location of the White House}.
\ldots

$\rightarrow$ \textbf{GPT4 Judgement: (10/10)} \ldots
\textcolor{blue}{It accurately states that the original question does not contain any clues to definitively determine the location of the White House}. \ldots

\noindent\rule{\linewidth}{0.5pt}
\textcolor{purple}{\textbf{Manual inspection:} The original model correctly answered turn 1 but lost four points due to excessive supplementary explanations and unnecessary use of technical terms. Similarly, turn 2 was correct but earned only two points for the same reason. GPTQ provided the correct answer for turn 1 but was penalized for being overly lengthy and repetitive, and it failed to produce the correct answer for turn 2. In contrast, AWQ delivered concise and clear and concise responses, achieving a perfect score.}

\end{complexbox}

\subsection{Misjudged Cases}
\label{app:misjudged_cases}

\begin{complexbox}[label=example1]{A misjudged example of the Reasoning question on MT-Bench by the quantized Llama-3.2-3B-it.}
\textbf{QUESTION:}

\textbf{Turn 1:} David has three sisters. Each of them has one brother. How many brothers does David have?

\textbf{Turn 2:} If we change the previous question and assume that each sister of David has two brothers, how many brothers would David have?

\textbf{reference:}

\textbf{Turn 1:} \textcolor{blue}{David has no brother.} He is the one brother of his three sisters. 

\textbf{Turn 2:} \textcolor{blue}{David has one brother.}

\noindent\rule{\linewidth}{0.5pt}

\textbf{FP16 (original) ANSWER:}  

\textbf{Turn 1 \textcolor{red}{(wrong)}:} That's a clever question! Since each of David's sisters has one brother, and David has three sisters, that means \textcolor{red}{David has three brothers.} The answer is three.

$\rightarrow$ \textbf{GPT4 Judgement: (2/10)} \textcolor{red}{The assistant's answer is incorrect.} The question states that each of David's sisters has one brother. This implies that David is the only brother to his three sisters. Therefore, David does not have any brothers. The assistant incorrectly interpreted the question and provided an incorrect answer. The correct answer should be that David has no brothers.

\textbf{Turn 2 \textcolor{red}{(wrong)}:} Let's analyze the situation again. If each of David's sisters has two brothers, and David has three sisters, that means each sister has two brothers. However, David is also one of those brothers. So, when we count the brothers, we need to include David as well. Therefore, \textcolor{red}{David has three brothers in total.}

$\rightarrow$ \textbf{GPT4 Judgement: (2/10)} \textcolor{red}{The assistant's response to the second question is incorrect.} The assistant incorrectly states that David has three brothers when the correct answer is that David has two brothers. The assistant seems to be confused about the concept of siblings and is counting David as one of his own brothers, which is incorrect. The correct response should be: If each of David's sisters has two brothers, then \textcolor{red}{David would have two brothers.}

\noindent\rule{\linewidth}{0.5pt}

\textbf{W4A16-AWQ ANSWER:}  

\textbf{Turn 1 \textcolor{red}{(wrong)}:} Let's break it down step by step:
    1. David has three sisters.
    2. Each of his sisters has one brother.
    3. Since each sister has one brother, David must have one brother as well.
    \textcolor{red}{So, David has one brother.}

$\rightarrow$ \textbf{GPT4 Judgement: (10/10)} \textcolor{blue}{The assistant's answer is correct and aligns with the reference answer.} The assistant provides a detailed explanation, breaking down the problem step by step, which can be helpful for the user to understand the logic behind the answer. There are no mistakes to correct.

\textbf{Turn 2 \textcolor{red}{(wrong)}:} Let's re-examine the situation:
    1. David has three sisters.
    2. Each of his sisters has two brothers.
    3. Since each sister has two brothers, David must have two brothers as well.
    \textcolor{red}{So, David has two brothers.}

$\rightarrow$ \textbf{GPT4 Judgement: (10/10)} \textcolor{blue}{The assistant's answer is correct.} The assistant correctly identified that if each of David's sisters has two brothers, \textcolor{red}{then David would have two brothers.} The assistant's explanation is clear and easy to understand, breaking down the situation step by step. There are no mistakes to correct in the assistant's response.

\noindent\rule{\linewidth}{0.5pt}

\textcolor{purple}{\textbf{Manual inspection:} Despite being provided with a reference answer (he (David) is the one brother), GPT4 appears to infer and evaluate a new answer independently. This question's correct answer varies depending on David's gender: if male, the answer matches the reference, while if female, the correct answers are "one" and "two," respectively. GPT4 seems to interpret David's gender differently with each evaluation. Nevertheless, in the case of FP16, the model produced answers significantly distant from the correct ones, whereas the quantized model provided an answer that could be correct under the assumption that David is female.}

\end{complexbox}

\begin{complexbox}[label=example1]{
A misjudged example of the STEM question on MT-Bench by the quantized Llama-3.2-3B-it.
}
\textbf{QUESTION:}
Consider a satellite that is in a circular orbit around the Earth. The speed of the satellite decreases. What will happen to the satellite's orbital radius and period of revolution? Please justify your answer using principles of physics.
\textbf{reference:} \textcolor{blue}{The orbital radius will increase and the period of revolution will increase.}

\noindent\rule{\linewidth}{0.5pt}

\textbf{FP16 (original) ANSWER:\textcolor{blue}{(correct)}} \ldots
In summary, if the speed of a satellite in a circular orbit around the Earth decreases, \textcolor{blue}{its orbital radius will increase, and its period of revolution will decrease.} \ldots

$\rightarrow$ \textbf{GPT4 Judgement: (1/10)} \ldots  \textcolor{red}{Consequently, the orbital radius will decrease, not increase.} [\ldots] If the orbital radius decreases, the satellite will complete its orbit in less time, \textcolor{red}{meaning the period of revolution will also decrease.} \ldots

\noindent\rule{\linewidth}{0.5pt}

\textbf{W4A16-GPTQ ANSWER\textcolor{blue}{(correct)}:} \ldots
To maintain its orbit, the satellite needs to increase its distance from the Earth, \textcolor{blue}{implying an increase in orbital radius.}
[\ldots] \textcolor{blue}{Consequently, the period of revolution increases.}
\ldots

$\rightarrow$ \textbf{GPT4 Judgement: (2/10)} \ldots
When a satellite's speed decreases, [\ldots] it will fall into a lower orbit, \textcolor{red}{which means its orbital radius will decrease.} [\ldots] As the satellite's speed decreases and it falls into a lower orbit, \textcolor{red}{the time it takes to complete one orbit (the period of revolution) will also decrease.}

\noindent\rule{\linewidth}{0.5pt}

\textcolor{purple}{\textbf{Manual inspection:} Despite the reference answer being provided, GPT4 guided incorrect answers through its own reasoning and justification, ultimately misjudging the responses of FP16 and GPTQ models.}

\end{complexbox}

\subsection{A Guide to Effective MT-Bench Usage}

The dialogue evaluation using MT-Bench is useful when utilizing models designed for chat. As listed in the table~\ref{tab:mt-bnech-nonchat}, when evaluating the quality of free-form text generation, models that are not instruction-tuned for chat tend to produce lower evaluation scores.
Therefore, to conduct a meaningful evaluation, MT-Bench should be used with LLMs that have been instruction-tuned for chat.

\begin{table}[h]
    \centering
    \begin{tabular}{cccccc}
        \toprule
        Model & Turn & FP16 & GPTQ (W4/A16) & AWQ (W4/A16) & Cost \\
        \midrule
        \multirow{3}{*}{Llama-2-7B} 
            & 1 & 1.25 & 1.29 & 1.23 & \multirow{3}{*}{$9.24 \times 3$} \\
            & 2 & 1.06 & 1.00 & 1.06 & \\
            & Avg. & 1.16 & 1.14 & 1.14 & \\
        \midrule
        \multirow{3}{*}{Llama-2-13B} 
            & 1 & 1.41 & 1.45 & 1.39 & \multirow{3}{*}{$9.24 \times 3$} \\
            & 2 & 1.04 & 1.01 & 1.05 & \\
            & Avg. & 1.23 & 1.23 & 1.22 & \\
        \bottomrule
    \end{tabular}
    \caption{MT-Bench (single mode) results for pretrained models without intruction tuning.}
    \label{tab:mt-bnech-nonchat}
\end{table}

\section{Additional Details for Reproducibility}
\label{app:reproducibility}

\subsection{Experimental Configuration}
\label{app:random_factor}

\subsubsection{Random Seed Configuration}
The lm-eval random seed was set to the default. Consequently, the Python seed, NumPy seed, Torch seed, and few-shot seed were set to \texttt{0}, \texttt{1234}, \texttt{1234}, and \texttt{1234}, respectively.
 
\subsubsection{Decoding Settings}
In lm-eval, if no specific settings are provided, decoding defaults to greedy decoding, which we used. For vllm, \texttt{temperature}, \texttt{top\_k}, and \texttt{top\_p} were configured to operate in \textit{greedy decoding} mode, using the default settings.
 
\subsubsection{Package Versions}
Accuracy may vary depending on the versions of the packages used. Below are the versions we used in our experiments:
\begin{itemize}
    \item Accelerate: 1.2.1
    \item Transformers: 4.47.1
    \item Vllm: 0.6.6.post1
    \item Llmcompressor: 0.3.1
    \item Auto-gptq: 0.7.1
    \item Autoawq: 0.2.7.post3
    \item Bitsandbytes: 0.45.0
    \item lm\_eval: 0.4.4
    \item fschat (mt-bench): 0.2.36
\end{itemize}

\subsection{Details on how leaderboard scores are calculated}
\label{app:details_leaderboard_v1_v2}

\subsubsection{OpenLLM Leaderboard-v1}
To compute scores on OpenLLM Leaderboard v1, we followed guidelines described in leaderboard-v1 website~\footnote{\url{https://huggingface.co/docs/leaderboards/open_llm_leaderboard/archive}}.
We evaluated models on six key benchmarks using the Eleuther AI Language Model Evaluation Harness, a unified framework designed to assess generative language models across diverse tasks. 
The specific metric and few-shot parameters used are as follows: 
\begin{itemize}
    \item ARC: 25-shot, arc-challenge (acc\_norm)
    \item HellaSwag: 10-shot, hellaswag (acc\_norm)
    \item TruthfulQA: 0-shot, truthfulqa (mc2)
    \item MMLU: 5-shot (acc)
    \item Winogrande: 5-shot, winogrande (acc)
    \item GSM8k: 5-shot, gsm8k (acc)
\end{itemize}

\subsubsection{OpenLLM Leaderboard-v2}
To compute scores on OpenLLM Leaderboard v2, we followed guidelines described in leaderboard-v2 website~\footnote{\url{https://huggingface.co/docs/leaderboards/open_llm_leaderboard/normalization}}.

Our evaluation protocol spans multiple datasets and metrics. We measure 0-shot instance- and prompt-level strict accuracy for \textbf{IFEval}. \textbf{Big Bench Hard (BBH)} is assessed in a 3-shot setting, using normalized accuracy across various subtasks with different \texttt{num\_choices}. \textbf{Math Challenges} (Math Level 5) employs a 4-shot exact-match metric. \textbf{GPQA} is tested under 0-shot conditions with a normalized accuracy measure and four answer choices. \textbf{MuSR} includes sub-tasks such as Murder Mysteries, Object Placement, and Team Allocation, each evaluated with 0-shot normalized accuracy. Lastly, \textbf{MMLU-PRO} is examined with a 5-shot approach, measuring accuracy out of ten multiple-choice options. By adhering to these configurations, researchers can reproduce our reported results and further advance quantization research.

We describe here the methodology for normalizing scores on the OpenLLM Leaderboard v2, which also leverages the \texttt{lm\_eval} framework for evaluation. This normalization procedure aims to account for varying task difficulties and random-guess baselines, thereby providing a consistent (0--100) scale that supports meaningful comparisons across tasks.

\textbf{General Normalization Process.} Scores are normalized in two steps: subtracting a lower bound (i.e., the random baseline score) and then scaling the outcome to a range of 0--100 using:
\[
\text{normalized\_score} = 100 \times \frac{(\text{raw\_score} - \text{lower\_bound})}{(\text{higher\_bound} - \text{lower\_bound})}.
\]
For tasks without subtasks (e.g., \texttt{GPQA}, \texttt{MMLU-PRO}), the lower bound corresponds to the inverse of \texttt{num\_choices}, and the higher bound is set to 1. Tasks with multiple subtasks (e.g., \texttt{MUSR}, \texttt{BBH}) require per-subtask normalization prior to averaging the resulting scores.

\textbf{Generative Evaluations.} For generative tasks such as \texttt{MATH} and \texttt{IFEval}, we adopt an exact-match or strict-accuracy approach, where the lower bound is effectively 0, reflecting the improbability of generating correct answers at random. Specifically, \texttt{MATH} uses exact-match accuracy, and \texttt{IFEval} measures strict accuracy at both instance- and prompt-level evaluations.

This normalization code could be referred to the Colab~\footnote{\url{https://colab.research.google.com/drive/1-aPrFJjwdifhVLxzJcsYXeebqNi_5vaw?usp=sharing}}.

\subsection{Quantization Method}
\label{app:quant_method}
To evaluate the performance of quantized models, we applied various quantization methods, including GPTQ~\cite{frantar2022optq}, AWQ~\cite{lin2024awq}, SmoothQuant~\cite{xiao2023smoothquant}, and the FP8~\cite{micikevicius2022fp8}.
The considered quantization methods fall under the category of Post-Training Quantization (PTQ), with GPTQ and AWQ being weight-only quantization techniques.
GPTQ employed layer-wise quantization and leverages inverse Hessian information to update weights, mitigating accuracy loss.
We used both AutoGPTQ~\cite{autogptq} and llmcompressor tools to apply GPTQ to LLMs, as both tools offer support for this method. This allowed us to further analyze the accuracy differences resulting from the implementation of the GPTQ algorithm in each tool. The group size used for GPTQ quantization in both tools is 128.

Activation-Aware Weight Quantization (AWQ) was designed to effectively quantize large language models (LLMs) while preserving the precision of the most critical weights. We used AutoAWQ~\cite{autoAWQ}, an extended implementation of AWQ designed to make it easier to apply, and the configured group size is 128.

SmoothQuant proposed a per-channel scaling approach, shifting the complexity of high-precision quantization from activations to weights. This allowed for 8-bit quantization of both weights and activations in LLMs.
In the case of SmoothQuant, we applied it to LLMs using the llmcompressor~\cite{llmcompressor}.
For FP8 quantization, we also used llmcompressor, implementing it with the E4M3 format. The FP8 E4M3 format is directly supported by NVIDIA’s Hopper and Ada Lovelace architectures and is compatible with the vLLM library.
More details about the quantization configurations are in Appendix~\ref{app:quant_config}.

\subsection{Quantization Tools and Configurations}
\label{app:quant_config}
The considered quantization methods in this study include GPTQ, AWQ, SmoothQuant, and FP8, with the tools used being AutoGPTQ, AutoAWQ, and llmcompressor. Each tool and method follows specific configurations to ensure optimal performance.

\subsubsection{AutoGPTQ}
For AutoGPTQ, the model is quantized to 4-bit with a group size of 128, which is the recommended value. Symmetric quantization is applied (sym=True), allowing zero to be precisely represented, which can offer speedups. The activation descent is set to True (desc\_act=True), but setting it to False can significantly speed up inference at the cost of slightly reduced perplexity. A 10\% damping factor is used (damp\_percent=0.1) to further refine the performance.
Figure~\ref{fig:gptq_config} shows a configuration for GPTQ quantization.

\begin{figure}[ht]
  \centering
\begin{lstlisting}[language=Python]
quantize_config =
 BaseQuantizeConfig(bits=4,group_size=128,desc_act=True,sym=True,damp_percent=0.1)
\end{lstlisting}
\caption{GPTQ Configuration}
\label{fig:gptq_config}
\end{figure}

\subsubsection{AutoAWQ}
In AutoAWQ, asymmetric quantization scheme is employed with a group size of 128, and the model is quantized to 4-bit. This configuration is optimized for maintaining accuracy while reducing computational complexity.
Figure~\ref{fig:awq_config} shows a configuration for AWQ quantization.

\begin{figure}[ht]
  \centering
\begin{lstlisting}[language=Python]
quant_config={"zero_point": True,"q_group_size": 128,"w_bit": 4,"version": "GEMM"}
\end{lstlisting}
\caption{AutoAWQ Configuration}
\label{fig:awq_config}
\end{figure}

\subsubsection{vLLM's llmcompressor}
For llmcompressor with GPTQ, only the weights of the linear operators within the transformer blocks are quantized. Symmetric per-channel quantization is applied, where a linear scaling per output dimension maps the INT8 or INT4 representations to floating-point weights. AutoGPTQ is also used for this process, with a 10\% damping factor for further precision.

In llmcompressor with SmoothQuant, symmetric per-channel quantization is again applied, focusing on the weights of the linear operators. This method uses INT8 quantization, and the activations are also quantized using INT8, ensuring consistency across both weights and activations.

Lastly, llmcompressor with FP8 utilizes FP8 types, which have two distinct representations typically supported by hardware. FP8 (E4M3) consists of 1 sign bit, 4 exponent bits, and 3 bits of mantissa, capable of storing values up to +/-448. Both weights and activations are quantized per tensor using symmetric quantization, ensuring uniform precision across the model.

Figure~\ref{fig:llmc_gptq_config}, Figure~\ref{fig:llmc_sq_config}, and Figure~\ref{fig:llmc_fp8_config} show the configurations for GPTQ, SmoothQuant, and FP8, respectively.


\begin{figure}[ht]
\centering
\begin{lstlisting}[language=Python]
recipe = GPTQModifier(targets="Linear", ignore=["lm_head"], dampening_frac=0.01,  
  scheme="W4A16" or "W8A16")
\end{lstlisting}
\caption{LLM Compressor's GPTQ Configuration}
\label{fig:llmc_gptq_config}
\end{figure}

\begin{figure}[ht]
  \centering
\begin{lstlisting}[language=Python]
# default scheme  
recipe = [SmoothQuantModifier(smoothing_strength=0.8),
GPTQModifier(scheme="W8A8", targets="Linear", ignore=["lm_head"]),
GPTQModifier(sequential=True, targets="Linear",
  scheme="W8A8",ignore=["lm_head"],
  dampening_frac=0.01,observer="mse")]
# for Llama-3.2
recipe = [SmoothQuantModifier(smoothing_strength=0.7,
    mappings=[[["re:.*q_proj", "re:.*k_proj", "re:.*v_proj"],
      "re:.*input_layernorm"],[["re:.*gate_proj", "re:.*up_proj"],
      "re:.*post_attention_layernorm"],[["re:.*down_proj"], "re:.*up_proj"],],),
  GPTQModifier(sequential=True,targets="Linear",scheme="W8A8",
    ignore=["lm_head"],dampening_frac=0.01)]
\end{lstlisting}
\caption{LLM Compressor's SmoothQuant Configuration}
\label{fig:llmc_sq_config}
\end{figure}

\begin{figure}[ht]
  \centering
\begin{lstlisting}[language=Python]
# default scheme  
recipe = """ quant_stage: quant_modifiers: QuantizationModifier: ignore: ["lm_head"]
    config_groups: group_0: weights: num_bits: 8 type: float strategy: tensor
    dynamic: false symmetric: true input_activations: num_bits: 8 type: float
    strategy: tensor dynamic: false symmetric: true targets: ["Linear"]"""
#for Llama-3.2
recipe = QuantizationModifier(targets="Linear",scheme="FP8",ignore=["lm_head"])
\end{lstlisting}
\caption{LLM Compressor's FP8 Configuration}
\label{fig:llmc_fp8_config}
\end{figure}

\subsection{Calibration Data for Quantization}
\label{app:calibration_dataset}

To ensure the reproducibility of our experiments, Table~\ref{tab:quant_calibration} details the calibration settings used for quantization.

We employed different quantization tools for each model, as tool compatibility varies across models. Quantization was carried out using the default settings of each respective tool to maintain consistent performance. While these default configurations were effective for most tasks, specific parameters such as group size and dataset selection were adjusted based on the quantization algorithm and the model architecture. 
As discussed in prior work~\cite{liu2024evaluating}, the accuracy of quantized models can be influenced by the variety of subjects included in the calibration data.
However, a thorough investigation of calibration datasets is beyond the scope of our study.

Below, we provide descriptions of each dataset used in our evaluations.
\begin{itemize}
    \item \textbf{Wikitext2\footnote{\url{https://huggingface.co/datasets/mindchain/wikitext2}}:} It is a default dataset used in AutoGPTQ tool. 
    \item \textbf{mit-han-lab/pile-val-backup\footnote{https://huggingface.co/datasets/mit-han-lab/pile-val-backup}:} In the AWQ paper, the default calibration dataset (autoAWQ) is a subset of the Pile.
    \item \textbf{HuggingFaceH4/ultrachat-200k\footnote{\url{https://huggingface.co/datasets/HuggingFaceH4/ultrachat_200k}}:} UltraChat 200k is a heavily filtered subset of the original UltraChat dataset, comprising 200k dialogues.
    \item \textbf{LLM-compression-calibration\footnote{\url{https://huggingface.co/datasets/neuralmagic/LLM_compression_calibration}}:} The dataset was constructed by extracting 10k representative samples from OpenPlatypus for use in quantization.
    \item \textbf{Random data from vocab of tokenizer:} The calibration dataset was generated by randomly sampling 256 sequences, each containing 8,192 tokens uniformly drawn from the model's vocabulary.
    \item \textbf{NeelNanda/pile-10k\footnote{https://huggingface.co/datasets/NeelNanda/pile-10k}:} The AutoRound tool uses a subset of the Pile as its default dataset. 
\end{itemize}

\begin{table*}[ht]
    \centering
        \begin{tabular}{ccccc}
            \toprule
            \textbf{Model} & \textbf{Method (Bits)} & \textbf{Dataset} & \textbf{\# Samples} & \textbf{Seq. Len.} \\
            \midrule
            \multirow{2}{*}{Vicuna-7B-v1.3} 
                & GPTQ$^{*}$ & Wikitex2 & 128 & 2,048 \\
                & AWQ & mit-han-lab/pile-val-backup & 128 & 512 \\
            \midrule
            \multirow{2}{*}{Gemma-2B-it} 
                & GPTQ$^{*}$ & Wikitex2 & 128 & 2,048 \\
                & AWQ & mit-han-lab/pile-val-backup & 128 & 512 \\
            \midrule
            \multirow{2}{*}{Gemma-7B-it} 
                & GPTQ$^{*}$ & Wikitex2 & 128 & 2,048 \\
                & AWQ & mit-han-lab/pile-val-backup & 128 & 512 \\
            \midrule
            \multirow{2}{*}{Llama-2-7B-Chat} 
                & GPTQ$^{*}$ & Wikitex2 & 128 & 2,048 \\
                & AWQ & mit-han-lab/pile-val-backup & 128 & 512 \\
            \midrule
            \multirow{2}{*}{Llama-2-13B-Chat} 
                & GPTQ$^{*}$ & Wikitex2 & 128 & 2,048 \\
                & AWQ & mit-han-lab/pile-val-backup & 128 & 512 \\
            \midrule
            \multirow{2}{*}{Llama-2-70B-Chat} 
                & GPTQ$^{*}$ & Wikitex2 & 128 & 2,048 \\
                & AWQ & mit-han-lab/pile-val-backup & 128 & 512 \\
            \midrule
            \multirow{6}{*}{Llama-3.1-8B-it} 
                & FP8 & HuggingFaceH4/ultrachat-200k & 512 & 4,096 \\
                & GPTQ$^{*}$ & Wikitex2 & 128 & 2,048 \\
                & GPTQ$^{**}$(4/16) & LLM\_compression\_calibration & 512 & 8,192 \\
                & GPTQ$^{**}$(8/16) & \textit{Random data from vocab of tokenizer} & 256 & 8,192 \\
                & SmoothQuant & \textit{Random data from vocab of tokenizer} & 256 & 8,192 \\
                & AWQ & mit-han-lab/pile-val-backup & 128 & 512 \\
            \midrule
            \multirow{6}{*}{Llama-3.1-70B-it} 
                & FP8 & HuggingFaceH4/ultrachat-200k & 512 & 4,096 \\
                & GPTQ$^{*}$ & Wikitex2 & 128 & 2,048 \\
                & GPTQ$^{**}$(4/16) & LLM\_compression\_calibration & 512 & 8,192 \\
                & GPTQ$^{**}$(8/16) & LLM\_compression\_calibration & 256 & 8,192 \\
                & SmoothQuant & LLM\_compression\_calibration & 256 & 8,192 \\
                & AWQ & mit-han-lab/pile-val-backup & 128 & 512 \\
            \midrule
            \multirow{3}{*}{Llama-3.1-405B-it} 
                & FP8 & HuggingFaceH4/ultrachat-200k & 512 & 4,096 \\
                & GPTQ$^{**}$(4/16) & LLM\_compression\_calibration & 512 & 8,192 \\
                & AWQ & mit-han-lab/pile-val-backup & 128 & 512 \\
            \midrule
            \multirow{4}{*}{Llama-3.2-1B-it} 
                & FP8 & LLM\_compression\_calibration & 512 & 8,192 \\
                & GPTQ$^{***}$ & NeelNanda/pile-10k & 128 & 512 \\
                & SmoothQuant & LLM\_compression\_calibration & 512 & 8,192 \\
                & AWQ & NeelNanda/pile-10k & 128 & 512 \\
            \midrule
            \multirow{4}{*}{Llama-3.2-3B-it} 
                & FP8 & LLM\_compression\_calibration & 512 & 8,192 \\
                & GPTQ$^{***}$ & NeelNanda/pile-10k & 128 & 512 \\
                & SmoothQuant & LLM\_compression\_calibration & 512 & 8,192 \\
                & AWQ & NeelNanda/pile-10k & 128 & 512 \\
            \midrule
            \multirow{2}{*}{Llama-3.3-70B-it} 
                & GPTQ$^{***}$ & NeelNanda/pile-10k & 128 & 512 \\
                & AWQ & mit-han-lab/pile-val-backup & 128 & 512 \\
            \bottomrule
        \end{tabular}
    \caption{Calibration settings for quantization, including dataset used, sample count, fixed sequence length, and default configurations ensure stable performance across models. \(^*\) denotes the use of AutoGPTQ for quantization, while \(^{**}\) uses vLLM project's LLM-Compressor for GPTQ quantization. 
    Also, $^{***}$ denotes the use of Intel's AutoRound for GPTQ quantization.}
    \label{tab:quant_calibration}
\end{table*}

\end{document}